\documentclass[11pt]{article}

\usepackage[preprint]{acl}

\usepackage{times}
\usepackage{latexsym}

\usepackage[T1]{fontenc}
\usepackage[utf8]{inputenc}
\usepackage{microtype}
\usepackage{inconsolata}

\usepackage{graphicx}
\usepackage{booktabs}
\usepackage{tabularx}
\usepackage{multirow}
\usepackage{makecell}
\usepackage{amsmath}
\usepackage{amssymb}
\usepackage[breakable, skins]{tcolorbox}
\usepackage{caption}
\usepackage{cleveref}
\usepackage{float}

\usepackage{xcolor}
\usepackage{tcolorbox}
\usepackage{multirow}
\usepackage{booktabs}
\usepackage{makecell}
\usepackage{colortbl}
\usepackage{subcaption}
\usepackage{enumitem}
\title{SpokenUS: A Spoken User Simulator for Task-Oriented Dialogue}

\author{Jonggeun Lee$^*$, Junseong Pyo$^*$, Jeongmin Park\textsuperscript{\S}, Yohan Jo$^{\dag}$ \\
Graduate School of Data Science, Seoul National University \\
\texttt{\{jonggeun.lee, junseongpyo, yohan.jo\}@snu.ac.kr}}

\begin{document}
\maketitle

\def\thefootnote{\fnsymbol{footnote}}
\footnotetext[1]{Equal contribution.}
\footnotetext[2]{Corresponding author.}
\footnotetext[4]{Visiting intern from Seoul National University.}
\def\thefootnote{\arabic{footnote}}

\begin{abstract}
Robust voice agents require exposure 
to the full diversity of how people interact through speech. However, obtaining enough spoken interactions is prohibitively expensive. Building spoken user simulators that address this requires large-scale spoken 
task-oriented dialogue (TOD) data encompassing spoken user behaviors, 
yet existing datasets are limited in scale and domain coverage, 
with no systematic pipeline for augmenting them. To address 
this, we introduce \textbf{SpokenTOD}, a spoken TOD dataset 
of 52,390 dialogues and 1,034 hours of speech augmented with 
four spoken user behaviors---cross-turn slots, barge-in, 
disfluency, and emotional prosody---across diverse speakers 
and domains. Building on SpokenTOD, we present \textbf{SpokenUS}, 
a spoken user simulator grounded in TOD that decides when to speak through a dedicated turn-taking head. SpokenUS achieves comparable goal 
coverage to much larger models while substantially 
outperforming all baselines in human MOS, disclosing slot values gradually across the dialogue as humans do rather than front-loading them. Further analysis confirms that SpokenUS's spoken behaviors pose meaningful challenges to voice agents, making it a practical tool 
for evaluating more robust spoken dialogue systems. Our code is available at \url{https://github.com/holi-lab/SpokenUS}.
\end{abstract}

\section{Introduction}

Human speech is rarely clean or scripted. People naturally produce disfluencies~\cite{shriberg1994preliminaries}, interrupt each other~\cite{sacks1974simplest}, and express emotion~\cite{schroder2001emotional}. These behaviors carry over to interactions with voice agents, where they are known to degrade agent performance~\cite{vocalbenchdf}. Building voice agents robust enough to handle this diversity is critical for real-world deployments like smart 
homes~\cite{seo2026simuhome} and on-device AI assistants~\cite{gunter2024appleintelligencefoundationlanguage}. These settings have long been studied as task-oriented dialogue (TOD), where systems must reliably help users accomplish concrete goals. Yet obtaining enough spoken interactions to train and evaluate such systems is prohibitively expensive.

\begin{table*}[t]
\centering
\footnotesize
\setlength{\tabcolsep}{5pt}
\renewcommand{\arraystretch}{1}

\begin{tabularx}{\textwidth}{
    p{0.12\textwidth}  
    p{0.14\textwidth}  
    X                  
    p{0.32\textwidth}  
}
\toprule
\textbf{Category} & \textbf{Type} & \textbf{Definition} & \textbf{Example or Details} \\
\midrule

\textbf{\mbox{Cross-Turn Slots}} &  & 
Slot values that span multiple turns. & 
Phone numbers, Email addresses. \\
 \midrule

\multirow{3}{*}{\textbf{Barge-in}} & Error Recovery & 
Correct an incorrect system response. & 
{\textbf{S:} ``For two peo...''
\textbf{U:} ``No, it's three.''} \\
\cmidrule{2-4}
& Clarification & 
Request details on an unclear response. & {\textbf{S:} ``It's nearby...''
\textbf{U:} ``Which address?''} \\
\cmidrule{2-4}
& Efficiency & 
Move on when sufficiently informed. & 
{\textbf{S:} ``Successfully...''
\textbf{U:} ``Thank you!''} \\
\midrule

\multirow{6}{*}{\textbf{Disfluency}} & Filled Pause & 
Non-lexical vocalized hesitation. & 
``[FP] uh, we should go there.'' \\
\cmidrule{2-4}
& Disc. Marker & 
Lexical filler with no semantic content. & 
``[DM] you know, the cheap one.'' \\
\cmidrule{2-4}
& Edit & 
Editing cue signaling a self-repair. & 
``[EDIT] I mean, somewhere cheaper.'' \\
\cmidrule{2-4}
& Restart & 
Abandoned utterance with a fresh start. & 
``book a ta- [RST] I'd like a table.'' \\
\cmidrule{2-4}
& Repetition & 
Immediate repetition without change. & 
``I want [REP] I want a single room.'' \\
\cmidrule{2-4}
& Correction & 
Substitution of a slot value. & 
``book it for Monday, [COR] Tuesday.'' \\
\midrule

\textbf{Emotion} & &\mbox{Emotional reactions of users.}  & 
Neutral, Satisfied, Dissatisfied, etc. \\

\bottomrule
\end{tabularx}
\caption{A taxonomy of spoken user behaviors in the SpokenTOD dataset. 
\textbf{S} denotes System and \textbf{U} denotes User.}
\label{tab:integrated_taxonomy}
\end{table*}

User simulation offers a scalable alternative \cite{658991}, where a user simulator is provided with a user goal (i.e., requests and all slot values) and interacts with an agent turn by turn to fulfill it. Yet recent LLM-based user simulators operate entirely in text \cite{llmusersimuatlor, shim2025noncollaborativeusersimulatorstool}. Attaching a Text-to-Speech module makes such a simulator audible and introduces spoken behaviors like emotional prosody, but it still receives the agent's utterance only once it is complete. It can therefore never cut in mid-utterance, leaving a core behavior of real users unmodeled and the agent's ability to recover from it untested.

Models that operate natively in speech offer an alternative, but they fall short in other ways. End-to-end omni models \citep{zeng2024glm, tong2025interactiveomni, xu2025qwen25omnitechnicalreport, xu2025qwen3omnitechnicalreport} process and generate speech directly, and full-duplex spoken dialogue models \citep{defossez2024moshi,  roy2026personaplex} go further by speaking and listening at once. Omni models, however, still provide no architectural support for the proactive turn-taking a user simulator must exhibit to test how assistants handle interruptions. Moreover, neither is grounded in TOD: they do not reliably convey the goal's slot values and requests across turns, and, having been optimized as helpful assistants, they default to help-providing rather than help-seeking behavior, at times drifting into the assistant role entirely (see Table~\ref{tab:personaplex_role_drift_example}).

Addressing these limitations is further complicated by data scarcity: existing spoken TOD datasets \cite{henderson-etal-2014-second, kim2021robust, si2023spokenwoz} are limited in scale and domain coverage, and no systematic pipeline exists for augmenting text-based TOD corpora with diverse spoken user behaviors, making it infeasible to train spoken user simulators grounded in both task-oriented domain knowledge and realistic spoken behaviors.

To address this gap, we introduce \textbf{SpokenTOD}, a large-scale spoken TOD dataset that augments existing text-based corpora with four spoken user behaviors: cross-turn slots, barge-in, disfluency, and emotional prosody (Table~\ref{tab:integrated_taxonomy}), spanning 52,390 dialogues and 1,034 hours of speech across diverse speakers and domains. Building on SpokenTOD, we present \textbf{SpokenUS}, a spoken user simulator grounded in TOD, supporting diverse spoken user behaviors, and deciding when to speak through a dedicated turn-taking head.

We evaluate SpokenUS against five end-to-end omni models across goal coverage, human MOS, text-speech alignment, and speaker similarity. SpokenUS achieves comparable goal coverage to substantially larger models, while disclosing slot values gradually across the dialogue as humans do, rather than front-loading them as omni models tend to do. It further outperforms all baselines in human MOS (4.06 vs.\ 3.18) despite being trained on orders of magnitude less audio data ($\sim$1K vs.\ $\sim$20M hours). We also demonstrate that SpokenUS's spoken behaviors pose a meaningful challenge to voice agents, yielding larger performance drops than other user simulators.
To our knowledge, SpokenUS is the first LLM-based spoken user simulator for evaluating (and potentially training) spoken dialogue systems, and we release it alongside SpokenTOD to support future work in this direction.

Our contributions are as follows:
\begin{itemize}
    \item We introduce \textbf{SpokenTOD}, a large-scale spoken TOD 
    dataset of 52,390 dialogues and 1,034 hours of speech, augmented 
    with cross-turn slots, barge-in, disfluency, and emotional
    prosody across diverse speakers and domains.
    \item We present \textbf{SpokenUS}, a spoken user simulator 
    grounded in TOD, supporting diverse 
    spoken user behaviors and deciding when to speak through a dedicated turn-taking head.
    \item We show that SpokenUS generates realistic spoken behaviors that pose a meaningful challenge to agents, making it a practical testbed for evaluating spoken dialogue systems.
\end{itemize}

\begin{figure*}[!t]
    \centering
    \includegraphics[width=1\textwidth]{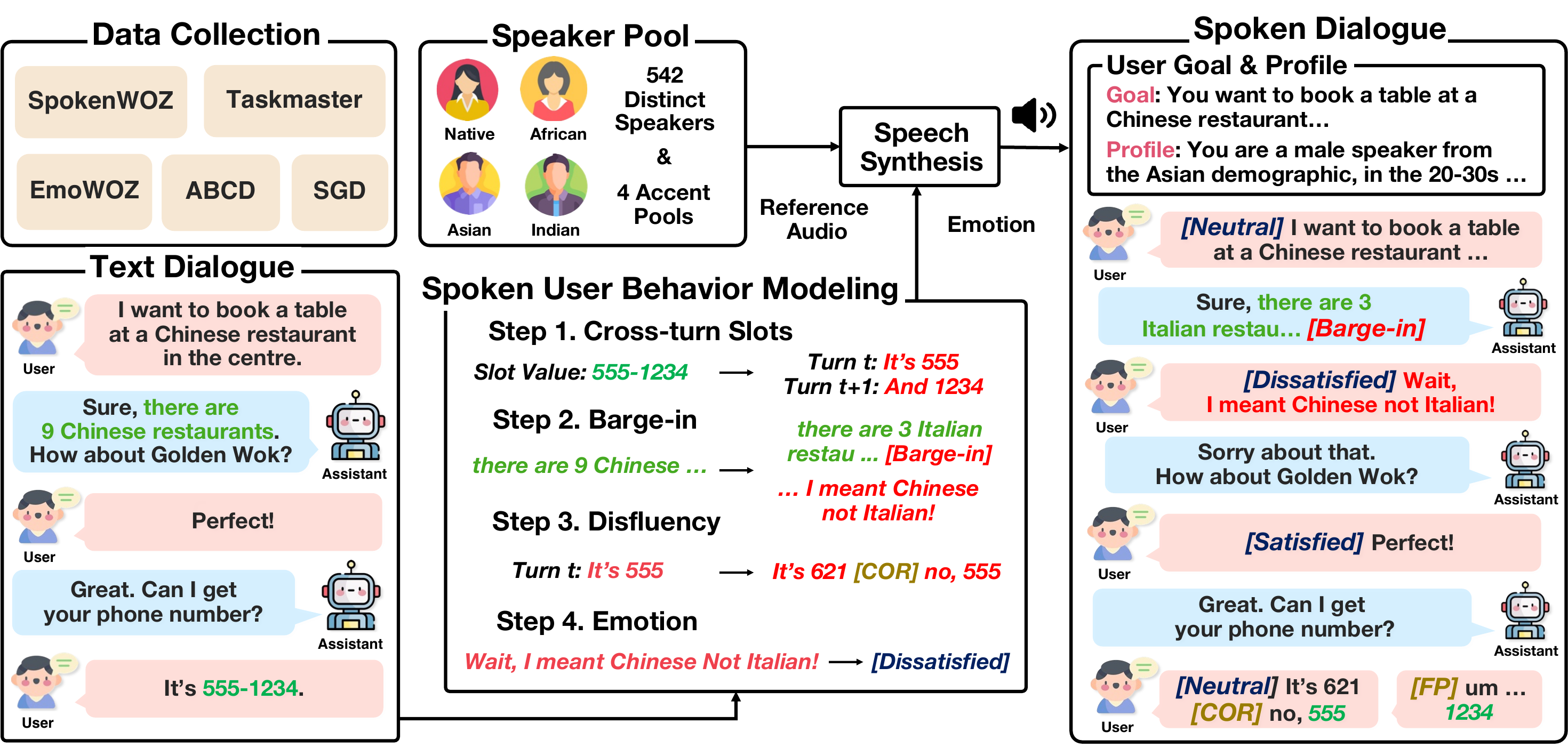}
    \caption{SpokenTOD Construction Pipeline.}
    \label{fig_overview}
\end{figure*}

\section{SpokenTOD}
To construct a spoken TOD dataset that captures realistic user behaviors, we augment existing text-based TOD datasets with these behaviors and synthesize the dialogues into speech rather than collecting spoken dialogues from scratch (Figure~\ref{fig_overview}). This approach offers two key advantages. First, by building on human-annotated corpora, we preserve both the conversational flow that emerges from real human interactions and rich annotations such as dialogue states and system acts. Second, our augmentation pipeline is dataset-agnostic: it can be layered on top of any existing TOD dataset construction pipeline, enabling generation of high-quality spoken TOD data across new domains.

We build upon four widely used datasets: ABCD \cite{chen2021action}, EmoWOZ (emotion-labeled MultiWOZ) \cite{feng2022emowoz}, SGD \cite{rastogi2020towards}, and Taskmaster \cite{byrne2019taskmaster}.

\subsection{Spoken User Behaviors}
In this section, we detail how we augment TOD datasets with spoken user behaviors: cross-turn slots, barge-in, disfluencies, and emotion (Table~\ref{tab:integrated_taxonomy}). For all LLM-based augmentation and annotation steps, we use Qwen3-32B \cite{yang2025qwen3}.

\paragraph{Cross-Turn Slots.}
\label{subsec:cross_turn_slot}
Humans often provide complex slot values such as phone numbers and email addresses across multiple turns rather than stating the full value at once \cite{MCINNES2004205, zhang-etal-2022-slot}, reflecting listeners' limited memory spans: each chunk is confirmed before the next is presented \cite{clark1991grounding}. Speakers may also revise previously provided slot values in subsequent turns. Recent work \cite{si2023spokenwoz} has identified these behaviors collectively as \emph{cross-turn slots}, a key challenge for voice agents that must collect, verify, and revise partial slot values across turns.
To model this, we segment all long alphanumeric slot values into natural chunks (Step 1 in Figure~\ref{fig_overview}) and additionally augment user-initiated self-corrections in a subset of them (e.g., Turn t: \textit{table for two} → Turn t+1: \textit{Actually, three}). Further details are provided in Appendix~\ref{cross_turn_augmentation_details}.

\paragraph{Barge-in.}
\label{subsec:turn_taking}
In natural spoken dialogue, speakers do not simply alternate complete utterances; they continuously monitor each other's speech and make real-time decisions about when to take or yield the floor \cite{sacks1974simplest, clark1996using}. We augment our dataset with \emph{barge-in}~\cite{khouzaimi2015turn}, where the user interrupts the agent to correct errors (\textit{error recovery}), request clarification (\textit{clarification}), or
skip ahead when sufficient information has been conveyed (\textit{efficiency}),
as summarized in Table~\ref{tab:integrated_taxonomy}. Since existing TOD datasets do not contain barge-ins, we modify the dialogues to insert natural barge-ins as follows. We first sample 25\% of user turns as barge-in candidates \cite{witt2015modeling}. After each candidate is judged contextually appropriate by the LLM, we insert a new agent utterance truncated mid-sentence, followed by the user's interruption. This barge-in is then naturally followed by the original agent response, which serves as a logical recovery response (Step 2 in Figure~\ref{fig_overview}). We validate the contextual appropriateness of these LLM-placed barge-ins through human evaluation, with 83\% rated as appropriate (Appendix~\ref{app:bargein_appropriateness}).
Detailed augmentation rules and prompt templates are provided in Appendix~\ref{turn_taking_augmentation_details}.

\begin{table*}[t]

\scriptsize
\centering
\setlength{\tabcolsep}{4pt}
\renewcommand{\arraystretch}{1.05}
\begin{minipage}[c]{0.42\textwidth}
    \centering
    \small
    \begin{tabular}{l r @{\hskip 0.2cm} l r}
    \toprule
    \multicolumn{2}{c}{\textbf{Dataset Overview}} & \multicolumn{2}{c}{\textbf{Behavior}} \\
    \midrule
    \# Dialogues  & 52{,}390      & \# Cross-turns   & 11{,}045 \\
    \# Utterances & 1,208,554 & \# Barge-ins     & 79{,}141 \\
    Avg.\ Words   & 9.95          & \# Disfluencies   & 118{,}377 \\
    \# Speakers   & 542           & \# Emotions    & 511{,}468 \\
    \textbf{Total Duration}  & \textbf{1{,}034 hrs}     &  \textbf{Total} & \textbf{720{,}031} \\
    \bottomrule
    \end{tabular}
\end{minipage}
\hspace{0.5cm}
\begin{minipage}[c]{0.54\textwidth}
    \vspace{0pt}
    \centering
    \includegraphics[width=\linewidth]{figs/fig_pie_chart.pdf}
\end{minipage}
\caption{Overall statistics of SpokenTOD.}
\label{tab:dataset_stats_combo2col}
\end{table*}

\paragraph{Disfluency.}
\label{subsec:disfluency}
Disfluency reflects the natural imperfections of real-time speech production, manifesting as hesitations, repetitions, and self-repairs \cite{shriberg1994preliminaries}. 
Our augmentation follows two stages. First, we select which utterances become disfluent using the length-dependent probability model of \citet{shriberg1996disfluencies}, where longer utterances are more likely to contain disfluencies. Second, we inject disfluencies from six categories, as shown in Table~\ref{tab:integrated_taxonomy}. Unlike previous work that uniformly inserts disfluencies via rules \cite{yang-etal-2020-planning,passali-etal-2022-lard} or LLM paraphrasing \cite{marie-2023-disfluency}, we concentrate insertion in a local window around slot values (Step 3 in Figure~\ref{fig_overview}), as speakers are known to be more disfluent when producing precise information such as names and phone numbers~\citep{CORLEY2007658,arnold2013information,sen-2020-speech}. Further details are provided in Appendix~\ref{disfluency_augmentation_details}.

\paragraph{Emotion.}
\label{subsec:emotion}
Beyond the behavioral patterns discussed above, realistic spoken dialogue also conveys speakers' emotional states. To support emotion-conditioned speech synthesis, we annotate each user utterance with an emotion label using the LLM (Step 4 in Figure~\ref{fig_overview}), following the EmoWOZ label set grounded in the Ortony--Clore--Collins (OCC) model \cite{ortony1988cognitive}: \textsc{Neutral}, \textsc{Fearful}, \textsc{Dissatisfied}, \textsc{Apologetic}, \textsc{Abusive}, \textsc{Excited}, and \textsc{Satisfied}. Agent utterances are all labeled as \textsc{Neutral}, reflecting the expectation that deployed AI systems communicate in a neutral manner \cite{han2023bots}. Annotation details and label reliability are in Appendix~\ref{emotion_annotation_details}.

\subsection{Speech Synthesis}
\label{subsec:synthesis}
We synthesize the augmented dialogues into speech using 
Qwen3-TTS \cite{hu2026qwen3}, controlling \emph{speech style} 
via emotion labels (\S\ref{subsec:emotion}) and \emph{speaker 
identity} via sampled audio from a speaker pool.

\paragraph{Emotion Conditioning.}
We control speech style in Qwen3-TTS via instruction prompts that specify the target emotion for each turn. Specifically, we map each emotion category to a set of speaking style keywords (e.g., \textsc{dissatisfied} $\rightarrow$ \{\textit{angry}, \textit{contempt}, \textit{disgusted}\}) drawn from CosyVoice3's training data \cite{du2025cosyvoice}, as both models share overlapping pre-training data. At each turn, we randomly sample one keyword from the corresponding set to introduce variation even within the same emotion label. Full construction details and the keyword list are provided in Appendix~\ref{emotion_conditioning_details}.

\begin{figure*}[!t]
    \centering
    \includegraphics[width=1\textwidth]{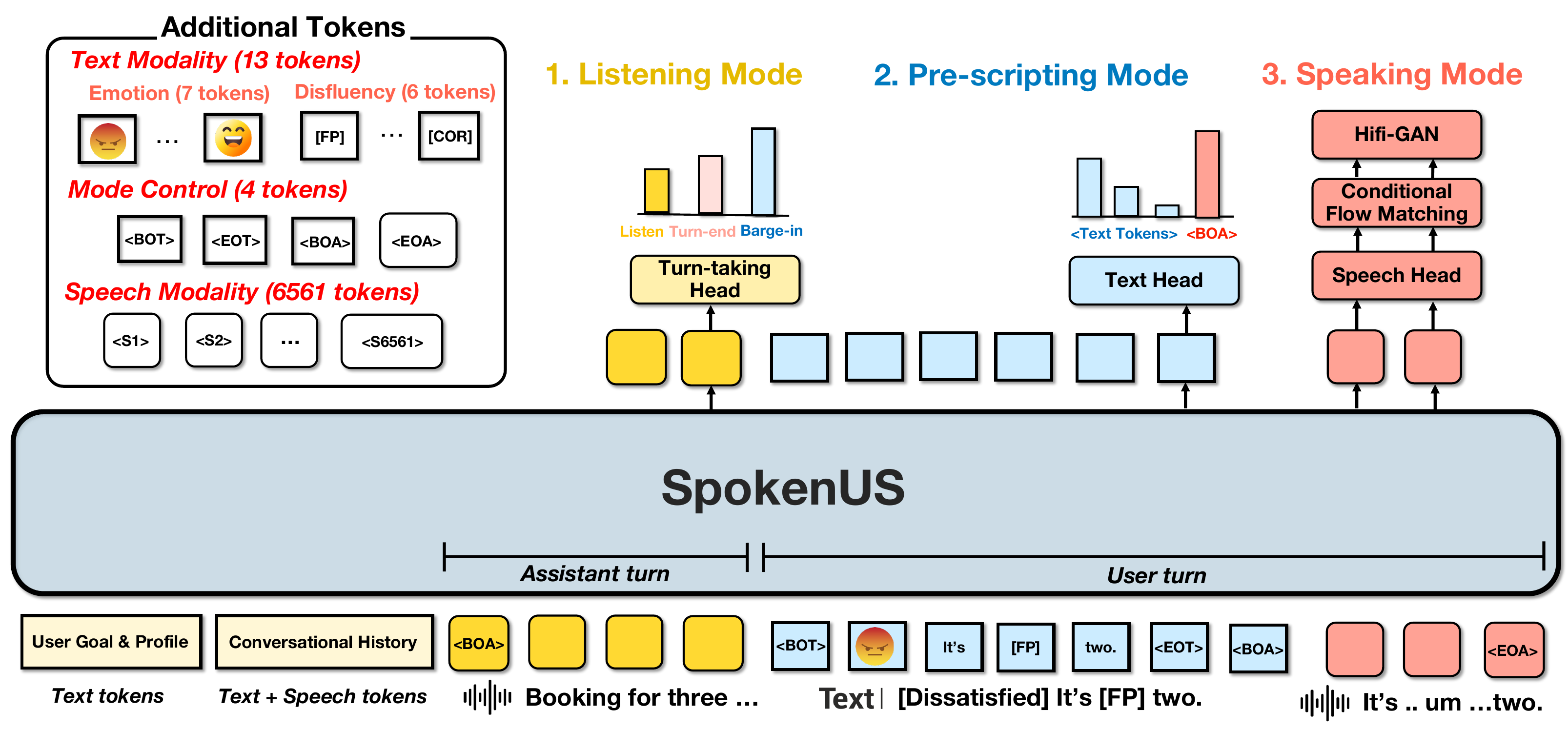}
    \caption{Overview of SpokenUS. The model processes streaming assistant speech in Listening Mode, determines barge-in timing through a turn-taking head, then generates responses through Pre-scripting and Speaking Modes.}
    \label{fig:spokenus_overview}
\end{figure*}

\paragraph{Speaker Diversity.}
To simulate a diverse range of real-world users, we sample reference audio clips from the Speech Accent Archive~\cite{weinberger2011speech}, weighting four accent groups by
population and stratifying by age and gender, yielding a pool of 542
reference speakers. For assistant utterances, we fix a pool of 10 \emph{Native}-accent speakers to reflect the consistent voice identity expected of deployed AI systems~\cite{cambre2019one}. Further details are in Appendix~\ref{audio_annotation_details}.

\subsection{Dataset Statistics}
\label{dataset_statistics}
To complement our synthesized data with authentic human speech, 
we additionally include SpokenWOZ~\cite{si2023spokenwoz} as 
human-recorded samples for both training and evaluation. The resulting SpokenTOD dataset comprises 52,390 dialogues and 1,034 hours of speech, split into train (39,430), validation (5,390), and test (7,570). As shown in Table~\ref{tab:dataset_stats_combo2col}, the augmentation pipeline produces 11,045 cross-turn slots, 79,141 barge-ins, 118,377 disfluencies, and 511,468 emotion annotations across 720,031 total behaviors, with the distribution of each behavior type illustrated in the pie charts.

\subsection{Dataset Validation}
We validate SpokenTOD along two axes: whether the synthesized speech is intelligible, and whether our augmentation contributes to realism beyond raw TTS synthesis. To measure intelligibility, we transcribe 500 sampled dialogues with Whisper-large-v3~\cite{radford2022robustspeechrecognitionlargescale} and compare against the source text, obtaining an overall WER of 4.69\%. To assess realism, three English-proficient annotators rate the original SpokenWOZ recordings (human recording), the same transcripts synthesized by Qwen3-TTS alone, and the full SpokenTOD pipeline on 5-point scales for voice naturalness and for how convincingly each spoken behavior is realized. Naturalness is comparable across all three settings (3.88–3.99), while augmentation markedly improves the realization of disfluency (2.97 → 3.76) and emotion (3.59 → 4.33). Further details are provided in Appendix~\ref{sec:dataset_validation}.

\section{SpokenUS}

\subsection{Framework Overview}

SpokenUS is a spoken user simulator that pursues a given user goal through multi-turn exchanges while expressing realistic spoken user behaviors. As illustrated in Figure~\ref{fig:spokenus_overview}, it takes the user's goal and profile, the conversational history as interleaved text and speech tokens, and the current assistant speech, provided either as a complete utterance or as a streaming input. SpokenUS operates in three sequential modes: \textit{Listening Mode}, which monitors incoming assistant speech to determine when to speak; \textit{Pre-scripting Mode}, which generates a transcript before speech is produced; and \textit{Speaking Mode}, which synthesizes the transcript into speech.

\subsection{Operating Modes}\label{operating_mode}

\paragraph{Listening Mode.}\label{sec:listening} The turn-taking head continuously 
monitors incoming assistant speech to determine when the user 
should speak. Since audio arrives as a stream of tokens~\cite{streaming}, 
the turn-taking head produces a prediction at each token and 
aggregates them by computing a weighted sum of per-token 
probabilities to avoid false alarms, with linearly increasing weights assigned to 
more recent tokens. A \textit{listen} 
prediction allows streaming to continue. A \textit{barge-in} prediction triggers an immediate interruption, while a \textit{turn-end} prediction indicates that the assistant has finished speaking. In either case, the model transitions to Pre-scripting Mode.
When a complete assistant utterance is provided as a single audio file rather than a stream, Listening Mode is skipped.

\paragraph{Pre-scripting Mode.} Upon entering this mode, the 
text head generates a structured transcript delimited by 
\texttt{<BOT>} and \texttt{<EOT>}. As illustrated in 
Figure~\ref{fig:spokenus_overview}, the transcript begins 
with an emotion token (e.g., \textit{[dissatisfied]}), 
followed by the intended utterance with interleaved 
disfluency tokens (e.g., \textit{[FP] two.}). This explicit 
transcript guides speech generation in Speaking Mode while 
making the simulator's behavior transparent for evaluation.

\paragraph{Speaking Mode.}\label{sec:speaking} The \texttt{<EOT>} token signals 
the transition to Speaking Mode, where the speech head 
autoregressively generates discrete speech tokens 
conditioned on the transcript, delimited by 
\texttt{<BOA>} and \texttt{<EOA>}. Since SpokenTOD is 
constructed with this same structure, emotion and disfluency 
markers are faithfully realized as prosodic and hesitation 
patterns. As illustrated in Figure~\ref{fig:spokenus_overview}, the generated tokens are decoded to audio via a Conditional 
Flow Matching model and HiFi-GAN~\cite{kong2020hifigan} 
vocoder, conditioned on a reference audio clip drawn from the speaker pool constructed in SpokenTOD 
(\S\ref{subsec:synthesis}), enabling SpokenUS to simulate 
a wide range of user voices.  Further details are provided in Appendix~\ref{app:operating_mode}.

\subsection{Training}\label{sec:training}
SpokenUS is trained with a multi-task objective combining losses from all three heads:
\begin{equation}
\mathcal{L} = \mathcal{L}_{\text{text}} + \mathcal{L}_{\text{speech}} + \mathcal{L}_{\text{turn-taking}}
\end{equation}
where $\mathcal{L}_{\text{text}}$ and $\mathcal{L}_{\text{speech}}$ are cross-entropy losses for text and 
speech token generation respectively, and $\mathcal{L}_{\text{turn-taking}}$ is the cross-entropy loss for 
turn-taking decisions over three classes (listen, barge-in, and turn-end). For turn-taking supervision, all tokens within an assistant turn are labeled \textit{listen} except for the final 6 tokens, which are labeled \textit{turn-end} or 
\textit{barge-in} depending on the turn type. The 6-token 
window corresponds to approximately 240 ms at 25 Hz, 
reflecting the average human reaction time when taking 
a turn~\cite{reaction_time}.  Speech 
token targets are obtained by encoding SpokenTOD audio with 
the CosyVoice3 tokenizer \cite{du2025cosyvoice}, which produces 
6,561 discrete tokens via Finite Scalar Quantization~\cite{fsq}.

\section{Experimental Setup}
\subsection{Implementation Details}
SpokenUS is initialized from Qwen2.5-3B 
\cite{qwen2025qwen25technicalreport} and fine-tuned in two stages. 
We first augment the vocabulary with mode control, disfluency, and 
emotion special tokens (Table~\ref{tab:special_tokens}). 
In Stage 1, the model is trained on text-only SpokenTOD to acquire TOD knowledge. In Stage 2, all three heads are jointly trained using the multi-task objective in Section~\ref{sec:training}, supervised on user turns only. 
The CFM and HiFi-GAN are also trained exclusively 
on user speech from SpokenTOD. Further training details are provided in Appendix~\ref{app:spokenus_training}.

\subsection{Baselines}
We compare SpokenUS against human recordings and five end-to-end omni models across two scales. Comparable-sized models are Qwen2.5-Omni-3B \cite{xu2025qwen25omnitechnicalreport} and InteractiveOmni-4B \cite{tong2025interactiveomni}; larger models are Qwen2.5-Omni-7B \cite{xu2025qwen25omnitechnicalreport}, GLM-4-Voice-9B \cite{zeng2024glm}, and Qwen3-Omni-30B-A3B \cite{xu2025qwen3omnitechnicalreport}. The assistant is a cascaded voice agent: user speech is transcribed by Whisper-large-v3~\cite{radford2022robustspeechrecognitionlargescale}, responses are generated by GPT-4.1-mini~\cite{openai_gpt4.1_2025}, and the reply is synthesized with Qwen3-TTS~\cite{hu2026qwen3}. The assistant is fixed across all simulators so that differences reflect the simulator alone. 

\subsection{Evaluation Metrics}
\label{sec:eval_metrics}
Unless otherwise noted, all evaluations are conducted on a fixed set of
100 goals taken from the SpokenWOZ test set (held out from training).
Every system is evaluated on this same set. SpokenWOZ provides the database states necessary for interactive user–assistant conversations, and its human recordings let us compare each simulator against a real speaker pursuing the same goal. Each system interacts with the assistant independently from the same goal, so systems differ in what is said as well as how it is said. 

\paragraph{Goal Coverage.}
We report \textit{Goal Alignment (GA)}, the percentage of dialogues in which \textit{all} required slot values are successfully conveyed, and \textit{Slot Match Rate (SMR)}, the percentage of required slots conveyed across all dialogues as a finer-grained complement. Both metrics are evaluated by GPT-4.1-mini~\cite{openai_gpt4.1_2025}, which has been shown to achieve high agreement with humans on slot verification in TOD~\cite{shim2025noncollaborativeusersimulatorstool}.

\paragraph{Human MOS.}
We recruit 10 English-proficient evaluators and randomly sample 20 dialogues from the evaluation set with 100\% GA. Each dialogue is rated by at least 3 evaluators on three 5-point scales: \textit{Naturalness (N)}, measuring voice quality; \textit{Conversational Flow (C)}, assessing role consistency as a help-seeker and naturalness of turn-by-turn engagement; and \textit{Spoken User Behavior (S)}, evaluating holistic authenticity of spoken behavior. Full details are provided in 
Appendix~\ref{app:human_eval}.

\paragraph{Text-Speech Alignment.}
We measure text-speech alignment via WER, computed between each model's text transcript and its corresponding synthesized speech using Whisper-large-v3~\citep{radford2022robustspeechrecognitionlargescale}. This captures how faithfully the intended text is realized in the generated speech, which is particularly important for omni models that generate text before synthesizing speech.

\paragraph{Speaker Similarity.}
We measure speaker consistency using cosine similarity between speaker embeddings extracted by wavlm-base-plus-sv~\cite{chen2022wavlm}. \textit{SIM}$(s_0, s_i)$ measures similarity between the first turn and each subsequent turn, reflecting how consistently the simulator maintains speaker identity throughout the dialogue. \textit{SIM}$(s_{i-1}, s_i)$ measures similarity between consecutive turns, capturing turn-level voice consistency.

\begin{table*}[t]
\centering
\small
\setlength{\tabcolsep}{5pt}
\begin{tabular}{lc|cc|cccc|c|cc}
\toprule
& & \multicolumn{2}{c|}{\textbf{Goal Cov.}} & \multicolumn{4}{c|}{\textbf{Human MOS}} & \textbf{Align.} & \multicolumn{2}{c}{\textbf{Speaker Sim.}} \\
\textbf{Model} & \textbf{Audio (hrs)} & GA $\uparrow$ & SMR $\uparrow$ & N $\uparrow$ & C $\uparrow$ & S $\uparrow$ & Avg. $\uparrow$ & WER $\downarrow$ & SIM$_{0,i}$ & SIM$_{i\text{-}1,i}$ \\
\midrule
Human & -- & 1.00 & 1.00 & 3.58 & 3.64 & 3.80 & 3.67 & -- & 0.84\tiny{$\pm$.07} & 0.88\tiny{$\pm$.04} \\
\midrule
\rowcolor{yellow!30} \multicolumn{11}{l}{\textit{Larger models}} \\
Qwen2.5-Omni-7B    & $\sim$3M    & 0.80          & 0.97          & 2.86          & 3.29          & \textbf{2.94} & 3.03          & \textbf{10.53} & 0.97\tiny{$\pm$.03} & 0.96\tiny{$\pm$.04} \\
GLM-4-Voice-9B     & $\sim$700K  & 0.33          & 0.79          & 2.51          & 2.06          & 2.38          & 2.32          & 15.33          & 0.94\tiny{$\pm$.04}          & 0.96\tiny{$\pm$.03}          \\
Qwen3-Omni-30B-A3B & $\sim$20M   & \textbf{0.93} & \textbf{0.99} & \textbf{3.03} & \textbf{3.61} & 2.91          & \textbf{3.18} & 15.05          & 0.95\tiny{$\pm$.03}          & 0.95\tiny{$\pm$.04}          \\
\midrule
\rowcolor{blue!20} \multicolumn{11}{l}{\textit{Comparable-sized models}} \\
Qwen2.5-Omni-3B    & $\sim$3M    & 0.36          & 0.93          & 1.97          & 2.77          & 2.28          & 2.34          & 15.09          & 0.90\tiny{$\pm$.04}          & 0.92\tiny{$\pm$.03}          \\
InteractiveOmni-4B & $\sim$465K  & 0.59          & 0.89          & 2.83          & 2.44          & 2.62          & 2.63          & 15.68          & 0.95\tiny{$\pm$.03}          & 0.94\tiny{$\pm$.04}          \\
SpokenUS (3B)      & $\sim$1K    & \textbf{0.82} & \textbf{0.99} & \textbf{3.92} & \textbf{4.14} & \textbf{4.11} & \textbf{4.06} & \textbf{11.36} & 0.93\tiny{$\pm$.02}          & 0.92\tiny{$\pm$.03}          \\
\bottomrule
\end{tabular}
\caption{Main evaluation results. 
The best result in each group is highlighted in \textbf{bold}.}
\label{tab:main_results}
\end{table*}

\section{Main Results}
Table~\ref{tab:main_results} presents results across all four metrics.

\paragraph{Goal Coverage.}
SpokenUS achieves a GA of 0.82, on par with Qwen2.5-Omni-7B (0.80) while 
substantially outperforming comparable-sized models 
(Qwen2.5-Omni-3B: 0.36; InteractiveOmni-4B: 0.59). Only 
Qwen3-Omni-30B-A3B surpasses it (0.93), at 10$\times$ the model size and $\sim$20M hours of training data, demonstrating that TOD grounding enables strong goal coverage at small scale. Failure cases are analyzed in Appendix~\ref{app:goal_alignment_analysis}. We further evaluate PersonaPlex~\cite{roy2026personaplex}, a full-duplex model with explicit role control, which reaches 0.06 GA: despite that control, it inverts the two roles, greeting the user and offering services rather than pursuing its goal (see Appendix~\ref{app:personaplex}).

\paragraph{Human MOS.}\label{result:human_mos}
SpokenUS achieves an average MOS of 4.06, substantially outperforming all omni-model baselines, including Qwen3-Omni-30B-A3B (3.18), despite being trained on orders of magnitude less audio ($\sim$1K vs.\ $\sim$20M hours). It is also the only model to score above the human recordings (3.67), with higher scores across all three criteria. Since each system independently generates a dialogue from the same user goal, this comparison reflects end-to-end interaction quality rather than the acoustic rendering of the same text (see Figure~\ref{fig:example_experiment_dialogue}). Human recordings otherwise outperform all omni model baselines; we discuss the interpretation of this result in Appendix~\ref{app:human_mos_details}.

\paragraph{Text-Speech Alignment.}\label{result:intelligibility}
SpokenUS achieves a WER of 11.36\%, comparable to 
Qwen2.5-Omni-7B (10.53\%) and substantially outperforming the 
remaining baselines (up to 15.68\%). Despite generating text 
before speech, omni models show poorer alignment, indicating that 
their synthesized speech often deviates from their own 
generated text.

\paragraph{Speaker Similarity.}
We interpret SIM relative to the natural range of variation observed in human speech, rather than as a measure where higher is better. Human recordings show relatively lower similarity scores (0.84 and 0.88), reflecting natural variation in emotion and prosody across turns. Omni models tend to produce overly uniform speech, yielding higher scores. SpokenUS closely approximates the human level (0.93 and 0.92), indicating that its explicit modeling of spoken behaviors introduces natural speaker variation. While Qwen2.5-Omni-3B shows similarly lower SIM scores (0.90 and 0.92), its significantly degraded speech quality (MOS: 2.34) suggests that this score does not reflect natural speaker diversity.

\section{Analysis}

\begin{figure}[t]
    \hspace{-0.2cm}
    \includegraphics[width=1\columnwidth]{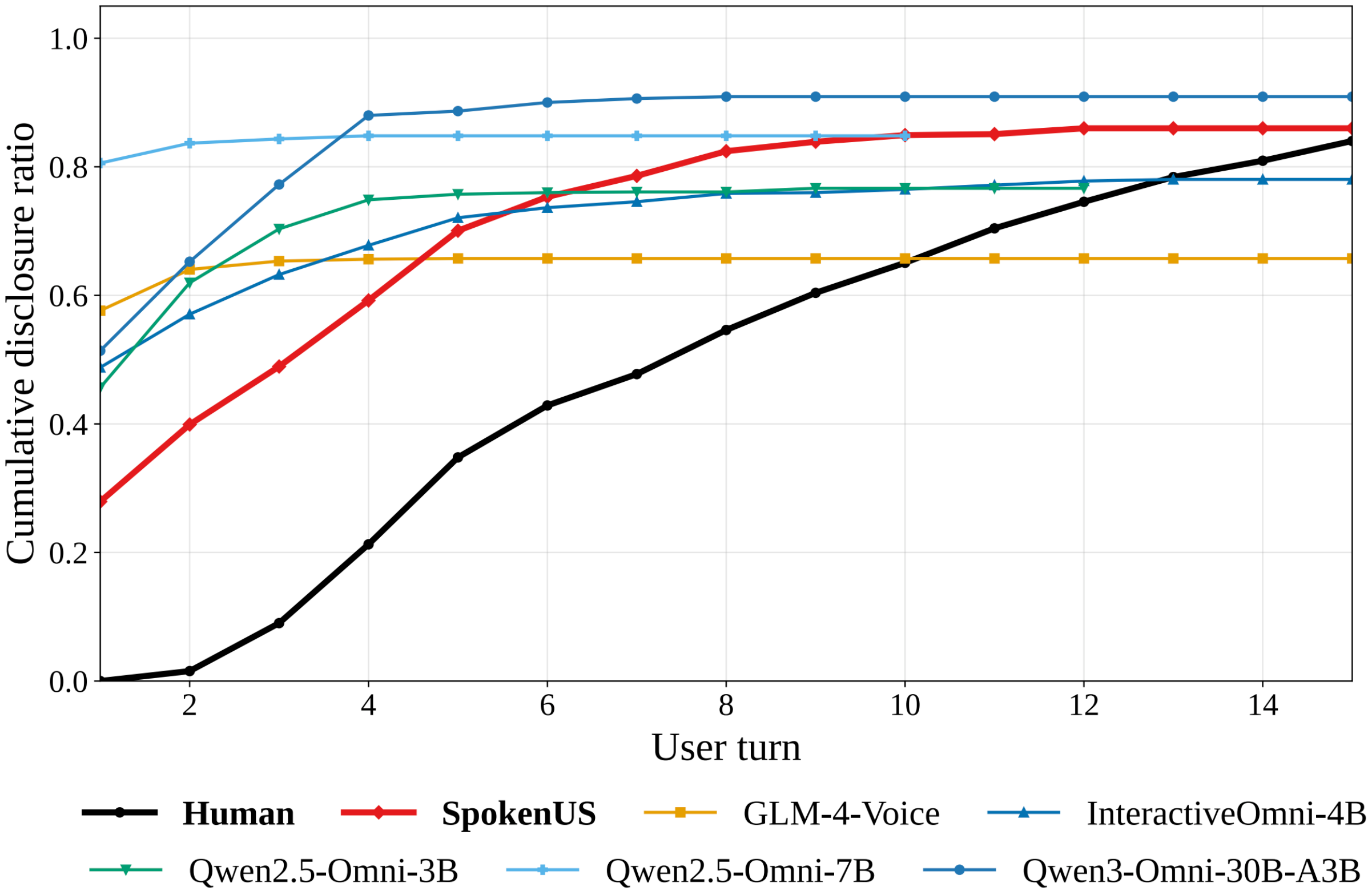}
    \caption{Cumulative goal slot coverage over user turns. }
    \label{fig:slot_timing}
\end{figure}

\subsection{Slot Disclosure Timing}\label{sec:slot_timing} 

Figure~\ref{fig:slot_timing} shows cumulative slot coverage over user turns: the proportion of goal slots that have appeared in the dialogue by each turn, normalized by the total number of slots per dialogue. We track only slot appearance rather than value correctness to isolate disclosure timing. Omni models tend to disclose slots within the first few turns and plateau early, whereas human speakers disclose them more gradually over the course of the dialogue. SpokenUS closely follows the human 
pattern, showing a steady increase in slot coverage throughout the dialogue rather than front-loading information. This pattern is not reproduced by instruction: prompting Qwen2.5-Omni-3B to disclose only one or two slots per turn leaves its curve almost unchanged, still saturating within the first few turns (see Appendix~\ref{app:disclosure_prompt}).

\subsection{Interruption Detection}

\begin{table}[t]
\centering
\small
\begin{tabular*}{\columnwidth}{@{\extracolsep{\fill}}lcccc}
\toprule
\textbf{Label} & \textbf{Correct} & \textbf{Early} & \textbf{Confused} & \textbf{Missed} \\
\midrule
Turn-end  & 66.0\% & 10.4\% & 16.4\% & 7.2\% \\
Barge-in  & 58.6\% & 18.0\% & 11.0\% & 12.4\% \\
\bottomrule
\end{tabular*}
\caption{Interruption performance on SpokenUS.}
\label{tab:interruption}
\end{table}

To evaluate whether SpokenUS correctly identifies when to speak during streaming assistant speech, we randomly sample 500 assistant turns labeled \textit{turn-end} and 500 labeled \textit{barge-in} from the SpokenTOD test set, processing each utterance sequentially from the beginning, as in Listening Mode (\S\ref{sec:listening}). Let the \textit{trigger window} denote the final 6 tokens of an assistant turn, where the supervision label is \textit{turn-end} or \textit{barge-in}. We categorize outcomes by where the model first fires a non-\textit{listen} prediction and which label it fires: \textbf{Correct} (within the window, correct label), \textbf{Confused} (within the window, wrong label), \textbf{Early} (fires before the window), and \textbf{Missed} (\textit{listen} throughout).

As shown in Table~\ref{tab:interruption}, turn-end detection achieves 66.0\% accuracy with a missed rate of 7.2\%, indicating that the model recognizes when the assistant has finished speaking. Barge-in detection is more challenging at 58.6\%, with a higher missed 
rate (12.4\%), reflecting the difficulty of deciding to intervene from partial context.
Notably, a substantial portion of errors stem from confusion between \textit{barge-in} and \textit{turn-end}. Since both labels lead the simulator to speak, this confusion is less consequential than a missed trigger: the model begins speaking within the trigger window in 82.4\% of turn-end and 69.6\% of barge-in cases, suggesting that the primary difficulty lies in disambiguating the two speaking behaviors rather than in detecting the moment to speak itself. Scoring and aggregation details are in Appendix~\ref{app:interruption}.

\subsection{Impact on Agent Performance}\label{sec:agent_impact}
We analyze how spoken user behaviors affect voice agent 
performance. To construct a controlled evaluation set, we allow each simulator up to 10 attempts per goal and retain the first dialogue that achieves 100\% GA, yielding 84 dialogues. This ensures F1 loss reflects the agent's tracking rather than information the user never conveyed. We evaluate Final Slot F1, the slot-level F1 between the agent's belief state at the final turn and the SpokenWOZ gold annotation, under three configurations: \textbf{Text}, where the agent receives the user transcript directly; \textbf{Cascaded}, where user speech is first transcribed via ASR; and \textbf{Audio}, where the agent receives speech directly, available only for the omni-modal agent.

\begin{table}[t]
\centering
\small
\setlength{\tabcolsep}{1.5pt}
\begin{tabular}{lccccc}
\toprule
& \multicolumn{2}{c}{\textbf{GPT-4.1-mini}} & \multicolumn{3}{c}{\textbf{Qwen2.5-Omni-7B}} \\
\cmidrule(lr){2-3}\cmidrule(lr){4-6}
\textbf{User Simulator} & Text & Cascaded & Text & Cascaded & Audio \\
\midrule
Qwen2.5-Omni-3B & 86.8 & 83.1\,{\scriptsize\textcolor{red}{$\downarrow$3.7}} & 62.5 & 54.1\,{\scriptsize\textcolor{red}{$\downarrow$8.4}} & 60.1\,{\scriptsize\textcolor{red}{$\downarrow$2.4}} \\
SpokenUS & 73.8 & 49.5\,{\scriptsize\textcolor{red}{$\downarrow$24.3}} & 62.0 & 48.3\,{\scriptsize\textcolor{red}{$\downarrow$13.7}} & 56.1\,{\scriptsize\textcolor{red}{$\downarrow$5.9}} \\
\bottomrule
\end{tabular}
\caption{Final Slot F1 across user simulators and agents.}
\label{tab:agent_robustness}
\end{table}

\begin{table}[t]
\centering
\small
\begin{tabular*}{\columnwidth}{@{\extracolsep{\fill}}lcc}
\toprule
\textbf{Assistant Model} & \textbf{Text} & \textbf{Cascaded} \\
\midrule
Qwen2.5-7B-Instruct & 73.5 & 57.2 {\scriptsize\textcolor{red}{$\downarrow$16.3}} \\
Llama-3.1-8B-Instruct & 65.8 & 45.7 {\scriptsize\textcolor{red}{$\downarrow$20.1}} \\
\bottomrule
\end{tabular*}
\caption{Final Slot F1 scores with SpokenUS on two additional agents.}
\label{tab:other_assistants}
\end{table}

As shown in Table~\ref{tab:agent_robustness}, SpokenUS already poses a greater challenge at the transcript level (73.8 vs.\ 86.8), indicating 
that human-like content patterns such as cross-turn slots and gradual slot disclosure are inherently harder for the agent to track.
The gap widens substantially in the Cascaded setting ($-$24.3 vs.\ $-$3.7), showing that spoken user behaviors themselves challenge the agent. This is not an artifact of poor synthesis quality, as evidenced by SpokenUS's low WER and high MOS (\S\ref{result:intelligibility}): even with an audio-capable agent that bypasses ASR, SpokenUS trails its transcript-level score by 5.9 points and remains harder than Qwen2.5-Omni-3B (56.1 vs.\ 60.1). This vulnerability is not specific to a single agent: two other assistant models show the same large cascaded drop on SpokenUS inputs ($-$16.3, $-$20.1; Table~\ref{tab:other_assistants}), indicating that the challenge stems from spoken behaviors rather than any one agent's weakness. Notably, error analysis reveals that agents tend to accept ASR's erroneous transcription of slot values rather than requesting confirmation, especially when those values are accompanied by disfluencies. Further details are provided in Appendix~\ref{app:agent_robustness_setup}.

\subsection{Generalization Beyond SpokenWOZ}
\label{app:sgd_eval}
To assess generalization beyond SpokenWOZ, we additionally evaluate user simulators on 100 dialogues from the SGD test set. Although SpokenTOD includes the SGD training split, the test split covers disjoint domains, allowing us to evaluate generalization to unseen domains. We use the PyTOD~\cite{coca2025pytod} framework with the same assistant and evaluation protocol as Section~\ref{sec:eval_metrics}.

\begin{table}[t]
\centering
\small
\begin{tabular*}{\columnwidth}{@{\extracolsep{\fill}}lcc@{}}
\toprule
\textbf{Model} & \textbf{GA $\uparrow$} & \textbf{SMR $\uparrow$} \\
\midrule
\rowcolor{yellow!30} \multicolumn{3}{l}{\textit{Larger models}} \\
Qwen2.5-Omni-7B    & 0.95 & 0.98 \\
GLM-4-Voice-9B     & 0.36 & 0.70 \\
Qwen3-Omni-30B-A3B & \textbf{0.98} & \textbf{0.99} \\
\midrule
\rowcolor{blue!20} \multicolumn{3}{l}{\textit{Comparable-sized models}} \\
Qwen2.5-Omni-3B    & 0.59 & 0.78 \\
InteractiveOmni-4B & 0.83 & 0.94 \\
SpokenUS (3B)      & \textbf{0.90} & \textbf{0.97} \\
\bottomrule
\end{tabular*}
\caption{Goal coverage and slot match rate on SGD.}
\label{tab:sgd_eval}
\end{table}

As shown in Table~\ref{tab:sgd_eval}, SpokenUS achieves 0.90 GA and 0.97 SMR, outperforming all comparable-sized baselines and remaining competitive with substantially larger models, consistent with the SpokenWOZ results. SpokenUS's goal coverage thus holds on a corpus with different domains from SpokenWOZ.

\section{Related Work}
\paragraph{User Simulation for TOD.}

User simulation has long been a cornerstone of TOD research, enabling agents to be trained under diverse conditions without the cost of collecting real interactions \citep{658991}. Early simulators \cite{schatzmann-etal-2007-agenda, chandramohan2012behavior} modeled user intent and dialogue acts via hand-crafted rules or learned policies. Recent studies \cite{llmusersimuatlor, shim2025noncollaborativeusersimulatorstool} have leveraged LLMs to generate more naturalistic user utterances, improving coverage and diversity. However, these simulators operate exclusively in text and do not account for the spoken behaviors, such as disfluencies \cite{shriberg1994preliminaries}, barge-ins \cite{khouzaimi2015turn}, and emotional prosody \cite{schroder2001emotional}, that emerge naturally in spoken interaction.

\paragraph{TOD Datasets.}
To support spoken TOD research, several spoken TOD datasets have been proposed. Early benchmarks such as DSTC2~\citep{henderson-etal-2014-second} and DSTC10~\citep{kim2021robust} evaluated TOD systems on spoken conversations but released only ASR transcripts and covered narrow domains. SpokenWOZ \cite{si2023spokenwoz} advanced the field with large-scale human-to-human recordings and fully annotated dialogue states. Nevertheless, human data collection remains costly and difficult to scale to new domains. Moreover, no existing work provides a systematic pipeline for augmenting text-based TOD corpora such as MultiWOZ \citep{budzianowski2018multiwoz}, SGD \citep{rastogi2020towards}, ABCD \citep{chen2021action}, and Taskmaster \citep{byrne2019taskmaster} with diverse spoken user behaviors across varied speakers and domains.

\paragraph{Spoken Dialogue Modeling.}
Early spoken dialogue systems relied on cascaded ASR--LLM--TTS pipelines~\cite{huang2024audiogpt}. Such pipelines suffer from compounding ASR errors~\cite{asrerror}, latency from sequential processing, and loss of non-linguistic cues such as emotional prosody and speaker characteristics~\cite{defossez2024moshi}. Generative spoken dialogue models \citep{nguyen2023generative, lee2025behavior} offer an alternative by modeling full conversations directly in the speech domain, but are designed to synthesize entire dialogues rather than engage interactively, making them unsuitable for turn-by-turn user simulation. End-to-end omni models \citep{xu2025qwen25omnitechnicalreport, xu2025qwen3omnitechnicalreport, zeng2024glm, tong2025interactiveomni} support interactive spoken dialogue, though without architectural support for proactive turn-taking. Full-duplex models such as Moshi \citep{defossez2024moshi}, LSLM \citep{ma2025language}, and PersonaPlex~\citep{roy2026personaplex} add this capability by speaking and listening at once. All of these models, however, are built for assistant-side generation rather than user-side simulation, and prompting them into the user role does not make them goal-faithful.

\section{Conclusion}
We introduced SpokenTOD and SpokenUS to address the lack of 
realistic spoken user simulation for task-oriented dialogue. 
SpokenTOD provides a large-scale spoken TOD dataset enriched with diverse user behaviors through an automated augmentation pipeline, and SpokenUS leverages it to train a spoken user simulator 
grounded in TOD knowledge with dedicated support 
for proactive turn-taking. Experimental results demonstrate that SpokenUS achieves strong 
goal coverage, while substantially outperforming all baselines in human MOS. 
Our agent robustness analysis further shows that SpokenUS's spoken 
behaviors pose a substantially greater challenge to voice
agents than existing omni models, underscoring the importance of 
realistic spoken user simulation for building and evaluating 
robust spoken dialogue systems.

\newpage

\section*{Limitations}
Both SpokenTOD and SpokenUS are currently limited to English, though the pipeline can be extended to other languages as long as spoken user behavior annotations and a TOD dataset in the target language are available. Additionally, SpokenUS models a predefined set of spoken behaviors; phenomena such as laughter and overlapping speech are not yet covered. Moreover, SpokenTOD relies on a single TTS backbone, which may not fully capture the acoustic variability of real-world environments, though a supplementary human evaluation finds its naturalness close to human recordings (3.88 vs.\ 3.95; Appendix~\ref{app:audio_quality}). Furthermore, SpokenTOD does not model adverse acoustic conditions commonly encountered in real-world deployments, such as background noise, reverberation, and low-bandwidth channels. Since the synthesized audio already exists as standalone files, such conditions can be introduced by simply applying standard acoustic augmentation (e.g., noise injection, room impulse response convolution) over the existing SpokenTOD speech. 

\section*{Acknowledgments}
This work was supported by the Ministry of Trade, Industry and Resources (MOTIR, Korea) and the Korea Planning \& Evaluation Institute of Industrial Technology (KEIT) (2410018399/25543163, Development of Contents and Software for AI Service Implementation on XR Smart Glasses). This work was also supported by the Institute of Information \& Communications Technology Planning \& Evaluation (IITP) grants funded by the Korean government (MSIT) (RS-2025-02215122, Development and Demonstration of Lightweight AI Model for Smart Homes; RS-2026-25505492, High-Performance Research AI Computing Infrastructure Support at the 2 PFLOPS Scale, contribution rate: 20\%), and by the IITP grant funded by the Korean government (MND) (No. RS-2026-25551943, Military-Industry-Academia Cooperation Center (Seoul (Yongsan))). This work was also supported by the National AI Project of the Ministry of Science and ICT and the AIIS.

\bibliography{custom}

\newpage
\appendix

\section{SpokenTOD Details}
\label{dataset_construction_details}
In this section, we provide a detailed breakdown of the data processing pipeline used to construct SpokenTOD.

\subsection{Dataset Details}
\paragraph{ABCD.}
Action-Based Conversations Dataset (ABCD) \cite{chen2021action} is a fully labeled human-to-human dataset containing over 10K customer service dialogues with 55 distinct user intents. Data were collected via an Expert Live Chat system pairing trained call center agents with crowdsourced users, with action labels annotated by agents in real time during the conversation.

\paragraph{EmoWOZ.}
EmoWOZ \cite{feng2022emowoz} is a large-scale emotion-annotated corpus built on MultiWOZ, containing over 11K dialogues with 83K user utterance-level emotion labels. Each utterance was annotated by three crowd workers via majority voting. The labelling scheme is grounded in the Ortony--Clore--Collins (OCC) model \cite{ortony1988cognitive}, defining seven emotion categories tailored to task-oriented settings: \textsc{Neutral},
\textsc{Fearful}, \textsc{Dissatisfied}, \textsc{Apologetic}, \textsc{Abusive}, \textsc{Excited}, and \textsc{Satisfied}. We adopt this label set for emotion annotation across all source datasets in SpokenTOD, using EmoWOZ's human-annotated labels directly where available.

\paragraph{SGD.} Schema-Guided Dialogue (SGD) dataset \cite{rastogi2020towards} contains over 20K task-oriented conversations spanning 20 domains such as travel, media, banking, and events. SGD adopts a schema-guided mechanism in which each service is described via natural language definitions of its intents and slots, rather than a fixed ontology. Dialogues were generated via a rule-based simulator and subsequently paraphrased by crowdsourced workers to improve naturalness, with slot annotations recovered automatically via string matching.

\paragraph{Taskmaster.}
We use Taskmaster-2 (TM-2), the second release of the Taskmaster corpus~\cite{byrne2019taskmaster}, which consists of 17K spoken two-person dialogues collected via a Wizard-of-Oz (WOZ) methodology across seven domains: restaurants, food ordering, movies, hotels, flights, music, and sports. User utterances were transcribed from audio recordings of crowdsourced workers via ASR. However, the original audio recordings are not publicly released.

\paragraph{SpokenWOZ.}
SpokenWOZ \cite{si2023spokenwoz} is the first large-scale speech-text benchmark for spoken TOD, comprising 5.7K dialogues, 203K turns, and 249 hours of audio from human-to-human spoken conversations across 8 domains. It introduces spoken-specific challenges such as cross-turn slots, which are absent in text-only datasets. As the only natively spoken source dataset in SpokenTOD, SpokenWOZ provides real acoustic and prosodic grounding for our dataset.

\paragraph{Dataset Licenses.}
All datasets are used in accordance with their respective licenses:
ABCD under the MIT License, Taskmaster-2 under CC BY 4.0, SpokenWOZ
and EmoWOZ under CC BY-NC 4.0, SGD under CC BY-SA 4.0, and the Speech
Accent Archive under CC BY-NC-SA 4.0. These datasets were originally
released for academic research in task-oriented dialogue systems, and our use is consistent with this
intended purpose.

\subsection{Dataset Unification}
\label{appendix:standardization}
To construct SpokenTOD from source datasets with varying formats, we unify each dataset into a common schema with two components: a user goal $G = (T, S)$, and slot value positions within each utterance for injecting disfluency near slot values (\S\ref{subsec:disfluency}). $G$ consists of a natural-language goal text $T$ and a structured goal $S$; $S$ records, for each sub-goal, the domain, intent type, and slot--value pairs (e.g., \texttt{find\_and\_book} in the \texttt{hotel} domain with \texttt{area: south, pricerange: moderate}).
Figure~\ref{fig:dataset_format} shows a simplified SpokenTOD dialogue in the resulting unified format.
As each source dataset has its own annotation format and characteristics, we apply dataset-specific preprocessing procedures as follows.

\paragraph{ABCD.} ABCD pairs each dialogue with a delexicalized version in which 
slot values are replaced with typed placeholders (e.g., \texttt{<username>}, 
\texttt{<email>}, \texttt{<order\_id>}). We recover slot value positions by 
aligning the delexicalized and original utterances via prefix/suffix matching; 
for values not covered by placeholders, we match them directly from the metadata. 
Since ABCD does not provide explicit user goals $G$, we reconstruct it from the 
scenario metadata: $S$ is constructed by mapping domain and intent labels with 
recovered slot--value pairs into the unified schema, from which $T$ is generated 
by filling a rule-based template with the customer prompt and scenario constraints.

\paragraph{EmoWOZ.} EmoWOZ extends MultiWOZ with per-turn emotion labels while 
inheriting its goal and slot annotations. We use the slot value positions from 
MultiWOZ directly. $G$ is taken directly from the MultiWOZ goal annotations: 
$S$ is mapped into the unified schema, and $T$ is the natural-language goal 
description provided by MultiWOZ. Emotion labels are used as provided.

\paragraph{SGD.} SGD annotates each turn with start and end positions for slot 
values, which we use directly. Since SGD does not provide explicit user goals 
$G$, we reconstruct it from turn-level annotations accumulated across the 
dialogue: $S$ is constructed by mapping turn-level intent and slot--value pairs 
into the unified schema, from which $T$ is generated by filling a rule-based 
template with the collected intents and constraints.

\paragraph{Taskmaster.} TM-2 annotates each turn with \texttt{start\_index} and 
\texttt{end\_index} fields marking slot value positions, which we use directly. 
Since TM-2 does not provide explicit user goals $G$ or intents, we reconstruct 
it from slot--value arguments collected across turns: $S$ is constructed by 
normalizing these into the unified schema, from which $T$ is generated by 
filling a rule-based template with the collected arguments.

\paragraph{SpokenWOZ.} SpokenWOZ provides dialogue state and dialogue act 
annotations from which we derive slot value positions. $G$ is constructed 
following the same procedure as EmoWOZ, using SpokenWOZ's goal annotations 
recorded in the MultiWOZ format.

\subsection{Cross-Turn Slots Details}
\label{cross_turn_augmentation_details}

Cross-turn slots represent slot values that span multiple turns rather than a single utterance, common with long alphanumeric sequences (e.g., phone numbers, email addresses, reservation IDs). We apply type-specific segmentation to all segmentable slot values: numeric sequences are split into 3--4 digit chunks, email addresses are decomposed into components with explicit punctuation vocalization (e.g., ``dot'', ``at''), and alphanumeric codes are split into letter and digit groups.

Speakers also extend a slot across turns by mis-speaking and repairing. We therefore inject mistakes with $P_{\text{error}} = 0.20$, following the error rate used in SpokenWOZ. This step is entirely rule-based: an incorrect segment is derived from the original by rule (e.g., altering digits of a phone number), and the resulting correction sub-dialogue---the user dictating the incorrect segment at turn $t$, self-correcting at turn $t+1$ (e.g., ``Wait, I meant...''), and the assistant acknowledging the correction---is filled from fixed templates with the corrected value.

\subsection{Barge-in Details}
\label{turn_taking_augmentation_details}
Barge-in refers to user interruptions during system speech, occurring
in up to 25\% of responses in spoken dialogue systems
\cite{witt2015modeling}. Following the turn-taking taxonomy of
\citet{khouzaimi2015turn}, we categorize barge-in by the assistant's
utterance state and the user's response type: (1) \textbf{Error
Recovery}: the user interrupts to correct inaccurate information in
the assistant's utterance;
(2) \textbf{Clarification}: the user interrupts due to comprehension
failure, expressing general confusion or requesting specific details;
(3) \textbf{Efficiency}: the user interrupts once enough information
has been conveyed to move the conversation forward. Each type is realized in one of three response
styles: \textbf{Implicit}, where the user
responds with minimal acknowledgment (e.g., ``Aha''); \textbf{Raw},
an explicit response that does not give the reason behind it (e.g., ``No, that's wrong''); and \textbf{Interpreted}, an explicit
response that states the reason or the user's own interpretation (e.g., ``No, I said 4 people, not 2''). Implicit pairs only with Efficiency, since a minimal
acknowledgment cannot convey a correction or a request for detail, yielding seven valid combinations.

We apply barge-in augmentation to 25\% of randomly sampled
user--assistant exchanges, uniformly assigning one of the seven
combinations to each. We prompt Qwen3-32B \cite{yang2025qwen3} to judge
whether the selected type can be naturally applied to the exchange and,
if so, to replace the assistant's response with its utterance truncated
at the interruption point (marked with a \texttt{\textless
bargein\textgreater} tag), the user's barge-in utterance in the selected
style, and the assistant's follow-up response. For Error Recovery, since
the original assistant utterances are correct, the model injects an
incorrect slot value into the truncated utterance, which the user then
corrects to the value in the current dialogue state. Prompts are provided in
Figures~\ref{fig:error_recovery_raw_prompt}--\ref{fig:efficiency_interp_prompt}.

\subsubsection{LLM Barge-in Appropriateness}
\label{app:bargein_appropriateness}
To verify whether the LLM places barge-ins at contextually 
appropriate moments during augmentation 
(\S\ref{turn_taking_augmentation_details}), we sample 20 barge-in 
turns from SpokenTOD and ask three annotators to judge whether each barge-in is contextually appropriate. A barge-in is judged 
appropriate when the assistant's preceding utterance contains 
a clear trigger for interruption, such as information 
conflicting with the user's stated goal (e.g., recommending 
a south-area restaurant when the user requested the centre). 
Across 60 total ratings (20 turns $\times$ 3 annotators), 
\textbf{83\%} of LLM-placed barge-ins are rated as contextually 
appropriate, indicating that the LLM understands the 
contextual conditions for barge-in: it constructs assistant turns whose 
preceding content naturally motivates user interruption. The 
audio-side realization of these barge-ins is further 
evaluated as part of the synthesized audio quality 
evaluation (Appendix~\ref{app:audio_quality}, Barge-in 
sub-dimension).

\subsection{Disfluency Details}
\label{disfluency_augmentation_details}

\paragraph{Disfluency placement.} 
We follow the annotation guidelines from the Switchboard Corpus~\citep{meteer1995dysfluency} for linguistically motivated disfluency placement. Filled pauses ([FP]), discourse markers ([DM]), and editing terms ([EDIT]) are sampled from fixed inventories and inserted before the target word. Repetitions ([REP]) duplicate the target span. These four types are applied by rule. Corrections ([COR]) replace the original slot value with an alternative, and restarts ([RST]) abandon the current utterance structure and begin a rephrased continuation; both require rewriting the utterance and are therefore generated by Qwen3-32B using the prompt templates in Figures~\ref{fig:disfluency-selfcorrection-prompt} and~\ref{fig:disfluency-restart-prompt}.

\paragraph{Scope and sampling.} We follow the exponential length-dependent model from~\citet{shriberg1996disfluencies}, where the probability of an utterance being disfluent increases with its length $L$ (measured in words):
\begin{equation}
P(\text{disfluent} \mid L) = 1 - b^L
\end{equation}
where $b$ represents the word-level fluency rate. Following~\citet{shriberg1996disfluencies}, we set $b = 0.9453$ based on the AMEX corpus \cite{kowtko1989data}, as it represents customer service interactions similar to task-oriented dialogue. This formulation ensures that longer utterances are more likely to contain disfluencies, reflecting patterns observed in spontaneous dialogue where speakers are more prone to self-corrections and hesitations in extended turns.

When an utterance is sampled to be disfluent, we uniformly sample one type from the six disfluency categories defined in Table~\ref{tab:integrated_taxonomy}: [FP], [DM], [EDIT], [REP], [COR], and [RST]. Following prior findings that disfluencies increase under higher planning and lexical retrieval difficulty~\citep{CORLEY2007658,arnold2013information,sen-2020-speech}, we inject them more frequently near slot values, as retrieving and articulating domain-specific entities demands greater cognitive effort. For all disfluency types except corrections, we insert them with 50\% probability near slots (within a 2-word window) and 50\% probability uniformly across the utterance, while corrections [COR] exclusively target slot values.

\subsection{Emotion Annotation Details}
\label{emotion_annotation_details}
To support emotion-conditioned speech synthesis, we annotate each user turn with seven emotion labels: \textsc{Neutral}, \textsc{Fearful}, \textsc{Dissatisfied}, \textsc{Apologetic}, \textsc{Abusive}, \textsc{Excited}, and \textsc{Satisfied}. We use Qwen3-32B \cite{yang2025qwen3} as our primary annotator. Annotation is skipped for EmoWOZ, as it already provides emotion labels. For turns that are cross-turn segments, we skip annotation and instead inherit the emotion label from the most recent non-segment user turn. The prompt is provided in Figure~\ref{fig:emotion_prompt}.

To ensure the reliability of LLM-based annotation, we validated Qwen3-32B against the ground truth labels of EmoWOZ. We sampled 50 instances per category (with 19 instances for Abusive due to its limited occurrence in the dataset, totaling 319 samples) and compared the model's predictions with the original labels. Qwen3-32B achieved an overall accuracy of 0.75, demonstrating its capability to capture affective states in task-oriented interactions. 

\subsection{Speech Synthesis Details}
\label{sec:synthesis_details}

Task-oriented dialogues frequently contain numbers, dates, times, addresses, and other entities that require normalization before speech synthesis. For instance, \textit{"I need a table for 2 at 7:30pm"} must be converted to \textit{"I need a table for two at seven thirty pm"} for proper pronunciation. 
We apply NeMo-text-processing \cite{zhang21ja_interspeech,bakhturina22_interspeech} 
to all utterances before synthesis. Key normalization patterns include:
\begin{itemize}
    \item \textbf{Numbers:} Cardinal (e.g., "123" → "one hundred twenty three") 
    and ordinal forms (e.g., "1st" → "first")
    \item \textbf{Dates and times:} "01/15/2024" → "January fifteenth twenty twenty four", 
    "7:30pm" → "seven thirty p m"
    \item \textbf{Currency and measurements:} "\$50" → "fifty dollars", 
    "5km" → "five kilometers"
    \item \textbf{Abbreviations and acronyms:} "Dr." → "doctor", 
    "NYC" → "N Y C" or "New York City" based on context
\end{itemize}

\begin{table}[t]
\centering
\small
\begin{tabularx}{\columnwidth}{@{}lX@{}}
\toprule
\textbf{Emotion Category} & \textbf{Style Keywords} \\
\midrule
\textsc{Neutral} & calm, indifferent, patient, relaxed \\
\textsc{Fearful} & fearful, shocked, surprised \\
\textsc{Dissatisfied} & angry, contempt, disgusted, defiant \\
\textsc{Apologetic} & compassionate, selfless, humble \\
\textsc{Abusive} & commanding, authoritative, merciless, loud, vengeful \\
\textsc{Excited} & adventurous, energetic, passionate, curious, creative, joyful \\
\textsc{Satisfied} & proud, hopeful, happy, cheerful \\
\bottomrule
\end{tabularx}
\caption{Emotion-to-keyword mapping for Qwen3-TTS instruction prompts. Keywords are drawn from the top-100 speaking styles in CosyVoice3's pre-training data.}
\label{tab:emotion_keywords}
\end{table}

\subsection{Emotion Conditioning}
\label{emotion_conditioning_details}
Qwen3-TTS conditions on emotion via instruction prompts of the form: \texttt{"Please speak in a \{emotion\} tone."} 

We construct our emotion keyword set from the top-100 speaking styles in CosyVoice3's pre-training data~\citep{du2025cosyvoice}. We first group these styles into seven coarse categories based on semantic similarity (neutral, fearful, dissatisfied, apologetic, abusive, excited, satisfied). We then evaluate the acoustic realization of each keyword through synthesis and select those that most effectively convey the intended emotion category, resulting in 3--6 keywords per category (Table~\ref{tab:emotion_keywords}). During synthesis, we randomly sample one keyword from the turn's assigned category to introduce prosodic variation while maintaining category consistency.

\subsection{Speaker Diversity}
\label{audio_annotation_details}
\paragraph{Speaker pool.} Reference audio clips from the Speech Accent Archive~\citep{weinberger2011speech} provide speaker identity and accent for zero-shot voice cloning. We target the accent distribution of the U.S.\ population (Table~\ref{tab:integrated_accent_data}), but the archive does not contain enough distinct speakers to realize this distribution with unique speakers alone. We therefore retain every qualifying speaker the archive provides in each accent group, keeping those whose reference clip is at most 25 seconds. This yields 542 unique speakers: 433 Native, 57 Asian, 38 Indian, and 14 African. For assistant utterances, we reserve a separate pool of 10 Native-accent speakers (5 male / 5 female), excluded from the user pool to avoid identity overlap.

\paragraph{Speaker sampling.} We recover the target distribution at sampling time by allowing speakers to recur across dialogues. A reference speaker is sampled in three stages: (1)~sample an accent pool using mixture weights derived from U.S.\ Census (2024) population estimates\footnote{\url{https://www.census.gov/data/datasets/time-series/demo/popest/2020s-national-detail.html}}, (2)~sample a country uniformly within the pool, and (3)~sample a speaker uniformly from that country. The accent distribution over utterances therefore follows the Census weights, with speakers in smaller pools recurring more often. We additionally stratify speaker age into four bins (10s, 20--30s, 40--50s, 60+) sampled with equal probability, and balance gender equally.

\begin{table}[t]
\centering
\footnotesize
\setlength{\tabcolsep}{3.5pt}
\begin{tabularx}{\columnwidth}{@{}>{\raggedright\arraybackslash}p{0.14\columnwidth}>{\raggedright\arraybackslash}X>{\raggedleft\arraybackslash}p{0.22\columnwidth}>{\raggedleft\arraybackslash}p{0.12\columnwidth}>{\raggedleft\arraybackslash}p{0.12\columnwidth}@{}}
\toprule
\textbf{Pool} & \textbf{Countries} & \textbf{Population} & \textbf{Weight} & \textbf{\# Spk.} \\
\midrule
Native  & USA, UK, Canada, Australia, Ireland      & 197,875,652 & 74.57\% & 433 \\
African & Ethiopia, Nigeria, Ghana, Senegal, Morocco & 42,951,595 & 16.19\% & 14  \\
Indian  & India, Pakistan, Afghanistan, Bangladesh, Nepal & 2,442,428 & 0.92\% & 38  \\
Asian   & China, South Korea, Japan, Philippines, Vietnam & 22,080,844 & 8.32\% & 57  \\
\midrule
\textbf{Total} & & \textbf{265,350,519} & \textbf{100.00\%} & \textbf{542} \\
\bottomrule
\end{tabularx}
\caption{Accent pool sampling weights based on U.S.\ Census (2024) estimates and resulting speaker counts.}
\label{tab:integrated_accent_data}
\end{table}

\subsection{Dataset Validation Details}
\label{sec:dataset_validation}

\subsubsection{Synthesized Audio Intelligibility}
\label{asr_quality}
To validate the intelligibility of our synthesized speech, we measure WER on 500 dialogues stratified by accent group using Whisper-large-v3~\citep{radford2022robustspeechrecognitionlargescale}. As shown in Table~\ref{tab:asr_wer}, our dataset achieves an overall WER of 4.69\%, confirming that the synthesized audio is sufficiently intelligible.

\begin{table}[t]
\centering
\small
\begin{tabular}{lcc}
\toprule
\textbf{Accent Group} & \textbf{WER (\%)} & \textbf{\# Utterances} \\
\midrule
African     & 5.08  & 1,340 \\
Asian                 & 3.77  & 1,242 \\
Indian                & 4.95  & 1,355 \\
Native       & 4.91  & 1,311 \\
\midrule
\textbf{Overall}      & \textbf{4.69}  & \textbf{5,248} \\
\bottomrule
\end{tabular}
\caption{ASR performance on synthesized user speech.}
\label{tab:asr_wer}
\end{table}

\subsubsection{Human Evaluation on SpokenTOD}
\label{app:audio_quality}
We conduct a supplementary human evaluation to assess (i)~how SpokenTOD
audio compares to authentic human recordings, and (ii)~whether our
augmentation pipeline contributes to spoken behavior realism beyond raw
TTS synthesis.

We compare three settings on 20 dialogues randomly
sampled from SpokenWOZ~\citep{si2023spokenwoz}: (i)~\textbf{Human}, the
original recordings serving as a ground-truth reference;
(ii)~\textbf{Qwen3-TTS}, which synthesizes the original transcripts with
Qwen3-TTS~\cite{hu2026qwen3} without our augmentation pipeline,
isolating raw TTS quality; and (iii)~\textbf{SpokenTOD}, the full
pipeline (augmentation $+$ Qwen3-TTS). For (ii) and (iii), the original
SpokenWOZ recording serves as the reference audio for voice cloning, so
speaker identity is held constant across the three settings and
differences reflect the pipeline rather than the voice.

Three annotators rate each setting on 5-point scales. All are
English-proficient speakers and compensated at local wage standards.
\textbf{Naturalness} asks how natural and human-like the user's voice
sounds, considering voice quality, rhythm, and intonation.
\textbf{Conversational Flow} asks whether the dialogue still proceeds
as a coherent exchange between a help-seeking user and an assistant. Four
\textbf{Spoken User Behavior} sub-dimensions then ask how convincingly
each augmented behavior is realized in speech.
Barge-in is marked N/A for Human and Qwen3-TTS, as the original
recordings contain no barge-in events and Qwen3-TTS synthesizes audio from 
the same transcripts.

\begin{table}[t]
\centering
\small
\setlength{\tabcolsep}{3pt}
\resizebox{\columnwidth}{!}{%
\begin{tabular}{lcc|cccc}
\toprule
& & & \multicolumn{4}{c}{\textbf{Spoken User Behavior}} \\
\cmidrule(lr){4-7}
\textbf{System} & \textbf{Nat.} & \textbf{Flow} & Cross & Barge & Disf. & Emo. \\
\midrule
\rowcolor{gray!15} \multicolumn{7}{l}{\textit{SpokenWOZ-sourced dialogues}} \\
Human       & 3.95 & 3.91 & 4.07 & N/A  & 3.96 & 4.56 \\
Qwen3-TTS   & 3.99 & 3.81 & 3.92 & N/A  & 2.97 & 3.59 \\
SpokenTOD   & 3.88 & 3.75 & 4.31 & 3.97 & 3.76 & 4.33 \\
\midrule
\rowcolor{gray!15} \multicolumn{7}{l}{\textit{Other source corpora}} \\
SpokenTOD   & 4.07 & 3.97 & 4.12 & 3.81 & 4.27 & 4.12 \\
\bottomrule
\end{tabular}%
}
\caption{Audio quality evaluation. The upper block compares three
settings on 20 SpokenWOZ-sourced dialogues; the lower block reports
SpokenTOD on 20 dialogues sampled from each of the four remaining
source corpora.}
\label{tab:audio_quality}
\end{table}

As shown in Table~\ref{tab:audio_quality}, SpokenTOD audio approaches
human-recorded quality in Naturalness (3.88 vs.\ 3.95). Conversational
Flow also stays close to the original (3.75 vs.\ 3.91), indicating that
inserting barge-ins preserves the coherence of the human-written
exchange. Comparing Qwen3-TTS without augmentation against the full
pipeline, augmentation improves Disfluency from 2.97 to 3.76
($+$0.79), Emotion from 3.59 to 4.33 ($+$0.74), and Cross-turn Slots
from 3.92 to 4.31 ($+$0.39). Since the two settings share the same TTS
backbone and reference voice, these gains are attributable to the
augmentation itself: without it, synthesized audio lacks the
disfluency and emotional variation that SpokenTOD explicitly models.

To test whether audio quality holds beyond SpokenWOZ-sourced
dialogues, we sample 20 dialogues from each of the four remaining
source corpora (SGD, Taskmaster-2, ABCD, EmoWOZ) and apply the same
protocol to the SpokenTOD setting. These corpora provide no original recordings, so only the
SpokenTOD setting is rated. Scores are comparable to or above the SpokenWOZ-sourced setting across all dimensions, indicating that
audio quality is consistent across the corpora used to construct
SpokenTOD. 

\section{SpokenUS Details}
\subsection{Operating Modes Details}\label{app:operating_mode}
\paragraph{Listening Mode.} Each assistant turn begins with the delimiter 
\texttt{<|im\_start|>}\allowbreak\texttt{assistant}\texttt{<BOA>}. SpokenUS processes each incoming speech 
token to obtain its last hidden state, which is passed to the turn-taking 
head. The head applies a softmax over three classes (listen, barge-in, and 
turn-end), and the resulting probabilities are aggregated via a linear-recency 
weighting scheme to determine whether to interrupt or yield.

\paragraph{Pre-scripting Mode.} Once a barge-in or turn-end is detected, a turn 
delimiter \allowbreak(\texttt{<|im\_end|>}\allowbreak\texttt{<|im\_start|>user<BOT>}) is inserted at 
that position. The last hidden state is then passed to the text head, which 
autoregressively generates the structured transcript until \texttt{<EOT>} is 
predicted.

\paragraph{Speaking Mode.} Upon predicting \texttt{<EOT>}, \texttt{<BOA>} is 
appended and the last hidden state of each subsequent token is routed to the 
speech head. Speech token generation continues until \texttt{<EOA>} is predicted, 
at which point the generated discrete speech tokens are converted to audio via the 
two-stage synthesis pipeline described in Section~\ref{sec:speaking}.

\subsection{SpokenUS Training Details}\label{app:spokenus_training}

\paragraph{SpokenUS.}
SpokenUS is initialized from Qwen2.5-3B~\citep{qwen2025qwen25technicalreport} 
and fine-tuned in two stages. We augment the vocabulary with disfluency and 
emotion special tokens, as listed in Table~\ref{tab:special_tokens}. In Stage 1, 
the model is trained on full text-only SpokenTOD dialogues (both user and assistant turns) 
for 3 epochs, enabling the model to acquire task-oriented dialogue knowledge 
such as goal-directed response generation. In Stage 2, all three heads are 
jointly trained for 12k steps using the multi-task objective in 
Section~\ref{sec:training}, supervised on user turns only. 
Hyperparameters are summarized in Table~\ref{tab:training_details}.

\begin{table}[t]
\centering
\small
\setlength{\tabcolsep}{4pt}
\begin{tabular}{ll}
\toprule
\textbf{Category} & \textbf{Tokens} \\
\midrule
Mode Control & \texttt{<BOT>}, \texttt{<EOT>}, \\
 & \texttt{<BOA>}, \texttt{<EOA>} \\
\midrule
Disfluency & \texttt{[COR]}, \texttt{[DM]}, \texttt{[EDIT]}, \\
 & \texttt{[FP]}, \texttt{[REP]}, \texttt{[RST]} \\
\midrule
Emotion & \texttt{<|abusive|>}, \texttt{<|apologetic|>}, \\ & \texttt{<|excited|>}, \texttt{<|satisfied|>}, \\ & \texttt{<|fearful|>}, \texttt{<|neutral|>}, \\ & \texttt{<|dissatisfied|>} \\
\bottomrule
\end{tabular}
\caption{Special tokens added to the \textbf{text} vocabulary: 4 mode-control, 6 disfluency, and 7 emotion tokens.}
\label{tab:special_tokens}
\end{table}

\begin{table}[t]
\centering
\small
\setlength{\tabcolsep}{8pt}
\begin{tabular}{lc}
\toprule
\textbf{Hyperparameter} & \textbf{Value} \\
\midrule
Backbone & Qwen2.5-3B \\
Hidden size & 2048 \\
Context length & 4,096 \\
\midrule
Optimizer & AdamW \\
Learning rate & $5 \times 10^{-5}$ \\
$\beta_1, \beta_2$ & 0.9, 0.999 \\
Weight decay & 0.01 \\
\midrule
Scheduler & Constant w/ warmup \\
Warmup steps & 2,500 \\
Gradient clipping & 1.0 \\
Gradient accumulation & 4 \\
Batch size & 10 \\
\midrule
Max epochs (Stage 1) & 3 \\
Training steps (Stage 2) & 12k \\
\midrule
Hardware & 5 $\times$ H200 \\
\bottomrule
\end{tabular}
\caption{Training hyperparameters for SpokenUS.}
\label{tab:training_details}
\end{table}

\paragraph{CFM and HiFi-GAN.}
The CFM and HiFi-GAN are both initialized from 
CosyVoice3~\citep{du2025cosyvoice} and fine-tuned exclusively 
on user speech from SpokenTOD. Training hyperparameters are 
summarized in Tables~\ref{tab:cfm_details} 
and~\ref{tab:hifigan_details}.

\begin{table}[h]
\centering
\small
\setlength{\tabcolsep}{8pt}
\begin{tabular}{lc}
\toprule
\textbf{Hyperparameter} & \textbf{Value} \\
\midrule
Parameters & 300M \\
DiT dim / depth / heads & 1024 / 22 / 16 \\
Mel dimension & 80 \\
Token frame rate & 25\,Hz \\
Token-mel ratio & 2 \\
Solver & Euler \\
$\sigma_{\min}$ & $1 \times 10^{-6}$ \\
CFG rate (train / infer) & 0.2 / 0.7 \\
\midrule
Optimizer & Adam \\
Learning rate & $1 \times 10^{-5}$ \\
Gradient clipping & 5.0 \\
Gradient accumulation & 4 \\
Warmup steps & 2,500 \\
Training steps & 30k \\
\bottomrule
\end{tabular}
\caption{Training hyperparameters for CFM.}
\label{tab:cfm_details}
\end{table}

\begin{table}[h]
\centering
\small
\setlength{\tabcolsep}{8pt}
\begin{tabular}{lc}
\toprule
\textbf{Hyperparameter} & \textbf{Value} \\
\midrule
Optimizer & Adam \\
Learning rate (G / D) & $1 \times 10^{-5}$ / $2 \times 10^{-4}$ \\
Gradient clipping & 5.0 \\
Gradient accumulation & 4 \\
Warmup steps & 2,500 \\
Discriminator warmup & 1,000 \\
Training steps & 30k \\
\midrule
Sampling rate & 24\,kHz \\
Mel dimension & 80 \\
Base channels & 512 \\
Upsample rates & {[8, 5, 3]} \\
\bottomrule
\end{tabular}
\caption{Training hyperparameters for HiFi-GAN.}
\label{tab:hifigan_details}
\end{table}

\section{Experiment Details}
\subsection{Evaluation Data}
\label{appendix:data_sources}
\begin{table}[t]
\centering
\small
\setlength{\tabcolsep}{6pt}
\renewcommand{\arraystretch}{1.1}
\begin{tabular*}{\columnwidth}{@{\extracolsep{\fill}} lc lc @{}}
\toprule
\multicolumn{2}{c}{\textbf{Slot Distribution (\%)}} & \multicolumn{2}{c}{\textbf{Domain Distribution (\%)}} \\
\midrule
Area         & 77\% & Train      & 50\% \\
Book\_people & 72\% & Restaurant      & 51\% \\
Book\_day     & 65\% & Attraction & 43\% \\
Phone    & 59\% & Hotel & 39\% \\
Departure    & 53\% & Taxi       & 19\% \\
Destination     & 53\% & Hospital   & 1\% \\
Day / Pricerange & 50\% & \\
\bottomrule
\end{tabular*}
\caption{Slot and domain distribution across 100 evaluation dialogues.}
\label{tab:test_set_stats}
\end{table}

We use the \textbf{SpokenWOZ test set} as our evaluation benchmark, as it provides human--human spoken recordings with fully annotated dialogue states and database queries necessary for running interactive user--assistant conversations. We sample 100 dialogues satisfying two criteria: (1) \textbf{domain coverage}, following the domain distribution of the source corpus, and (2) \textbf{human goal alignment}, retaining only dialogues in which the human user successfully conveys all required goal slots (GA $= 1.0$). The latter ensures that any gap in goal alignment across simulators reflects simulator capability rather than inherent dialogue difficulty. Table~\ref{tab:test_set_stats} summarizes the resulting distribution. \textit{Hospital} is rare in the source corpus and rarer still after the second criterion, so we include all dialogues available in this domain.

\subsection{Baseline Inference}
\label{app:baseline_inference}

All baseline models except InteractiveOmni-4B are queried with 
temperature $= 0$ and top-$p = 1.0$. InteractiveOmni-4B follows 
the official configuration (\texttt{do\_sample=True}, 
\texttt{max\_new\_tokens=1024}), with temperature and top-$p$ 
at their default values.

\subsection{Goal Alignment Details}
\label{app:goal_alignment}
Goal alignment is evaluated turn-by-turn using GPT-4.1-mini with temperature $= 0$. At each user turn, the model is presented with the remaining uncovered goal items and the dialogue history, and selects which items are explicitly mentioned, confirmed, or requested in the user's utterance. A dialogue is considered goal-aligned if all goal items are covered by the end of the conversation. \textbf{Goal Alignment (GA)} is the percentage of dialogues that are fully goal-aligned, and \textbf{Slot Match Rate (SMR)} is the fraction of slot-type goal items covered, micro-averaged across all dialogues. Both metrics are computed from a single run with fixed inference parameters. This evaluation protocol follows \citet{shim2025noncollaborativeusersimulatorstool}, who report an MCC of 0.77 between LLM and human annotations on a comparable goal alignment task, supporting the reliability of this approach. The prompt used for goal alignment evaluation is provided in Figure~\ref{fig:goal_align_prompt}.

\subsection{Goal Alignment Analysis}
\label{app:goal_alignment_analysis}
We perform error analysis on dialogues in which each model fails to convey all required goal slots, from the 100-dialogue SpokenWOZ test subset used in Table~\ref{tab:main_results}. Since each SpokenWOZ goal typically spans multiple domains (e.g., booking a restaurant and arranging a taxi), we categorize each error both by its content type and by whether it occurs in the initial domain or in a later domain after a domain shift.

\paragraph{Error Taxonomy.}
Recall that each structured goal $S$ records a set of sub-goals, each consisting
of a \textit{domain}, an \textit{intent}, and associated slot--value pairs
(\S\ref{appendix:standardization}). For error analysis, we partition these
slot--value pairs into two types. \textit{Constraints} are conditions required to
find or book the target, such as area, price range, or day. \textit{Requests} are
additional information the user should ask for after the target has been found,
such as postcode, phone number, or car type. For example, the goal \textit{``Find
an expensive Indian restaurant in the west and ask for its postcode and phone
number''} yields \textit{constraints} = \{food=Indian, area=west,
pricerange=expensive\} and \textit{requests} = \{postcode, phone\}. Based on this decomposition, we define three dialogue-level error types: \textbf{Constraint Omission}, where only constraints are missed; \textbf{Request Omission}, where only requests are missed; and \textbf{Composite Omission}, where both occur within the same dialogue.

\begin{table}[t]
\centering
\small
\setlength{\tabcolsep}{5pt}
\renewcommand{\arraystretch}{1.1}
\begin{tabular*}{\columnwidth}{@{\extracolsep{\fill}} lccc @{}}
\toprule
\textbf{Omission} & \textbf{Initial (\%)} & \textbf{Later (\%)} & \textbf{Total (\%)} \\
\midrule
Constraint & 4.5  & 13.6 & 18.2 \\
Request    & 22.7 & 59.1 & 81.8 \\
Composite  & 0.0  & 0.0  & 0.0  \\
\bottomrule
\end{tabular*}
\caption{Slot-level error distribution of SpokenUS by omission type and dialogue position. \textit{Initial} denotes the first domain in the realized dialogue flow; \textit{Later} denotes subsequent domains after a domain transition.}
\label{tab:spokenus_error_type_by_position}
\end{table}

\begin{table}[t]
\centering
\resizebox{\columnwidth}{!}{%
\begin{tabular}{lccc}
\toprule
\textbf{Model} & \textbf{Constraint (\%)} & \textbf{Request (\%)} & \textbf{Composite (\%)} \\
\midrule
Qwen2.5-Omni-7B & 35.0 & 65.0 & 0.0 \\
Qwen3-Omni-30B-A3B  & 42.9 & 28.6 & 28.6 \\
Qwen2.5-Omni-3B & 28.1 & 54.7 & 17.2 \\
GLM-4-Voice-9B    & 25.4 & 35.8 & 38.8 \\
InteractiveOmni-4B   & 31.7 & 36.6 & 31.7 \\
SpokenUS        & 22.2 & 77.8 & 0.0 \\
\bottomrule
\end{tabular}%
}
\caption{Dialogue-level omission-type distribution across models. SpokenUS exhibits a narrower, request-heavy failure profile compared to the other models.}
\label{tab:model_error_type_distribution}
\end{table}

As shown in Table~\ref{tab:spokenus_error_type_by_position}, although SpokenUS achieves a high overall goal-alignment rate, its remaining failures at the slot level are highly concentrated in request omission and later-domain subsets. This suggests that SpokenUS generally preserves the core goal structure, but is less reliable in completing follow-up requests after a domain transition. At the dialogue level (Table~\ref{tab:model_error_type_distribution}), the other models exhibit lower goal-alignment rates and broader error profiles. Their failures are not concentrated in request omission, but are more evenly distributed across constraint, request, and composite omissions, indicating more diffuse weaknesses in overall goal realization. This contrast suggests that future improvements to SpokenUS should focus on not only increasing goal alignment, but also tracking unresolved follow-up requests before proceeding to the next domain.

\subsection{Prompting for Gradual Disclosure}
\label{app:disclosure_prompt}

\begin{figure}[t]
    \hspace{-0.2cm}
    \includegraphics[width=1\columnwidth]{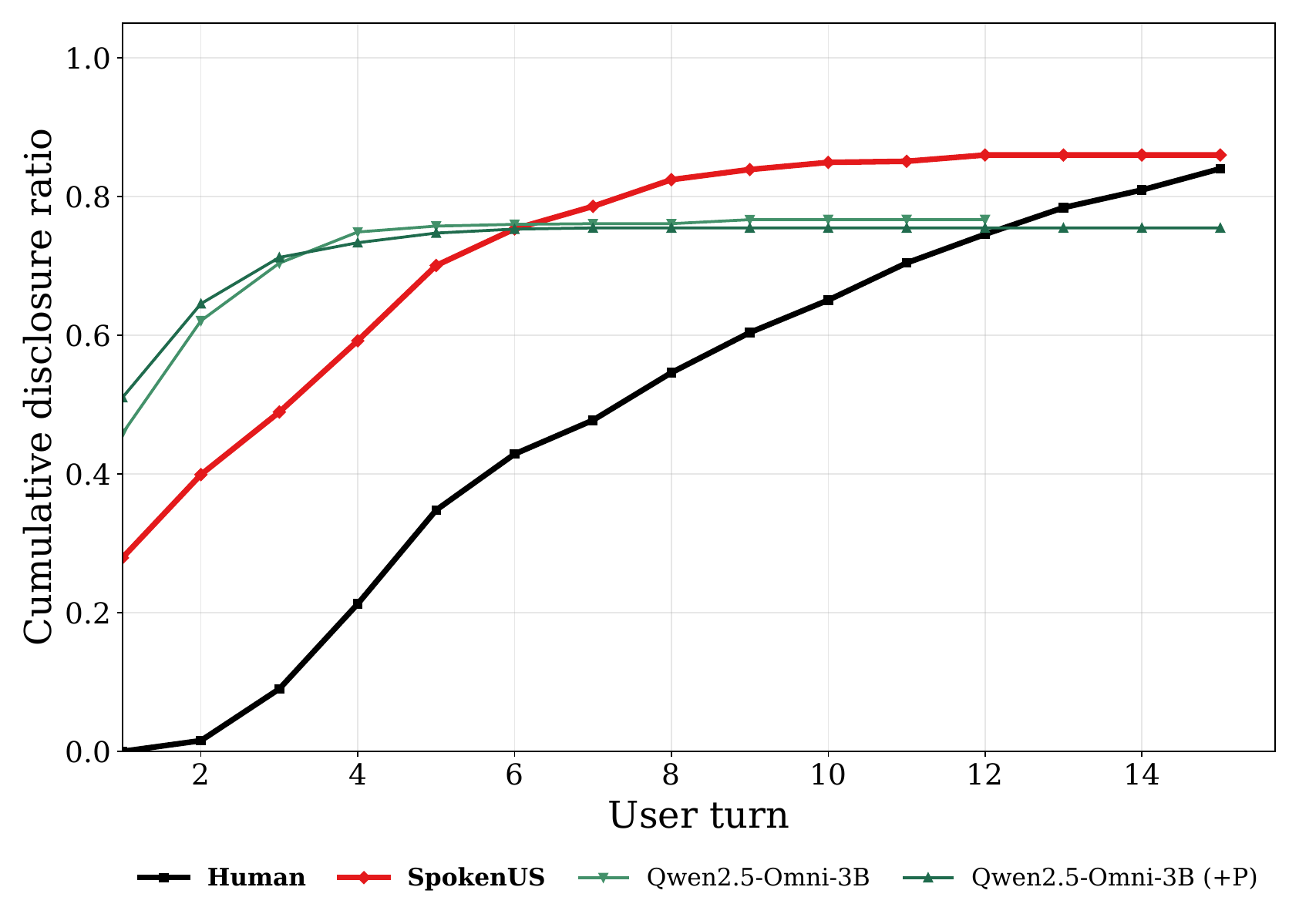}
    \caption{Cumulative goal slot coverage over user turns with and without gradual-disclosure prompting (\texttt{+P}).}
    \label{fig:slot_timing_with_p}
\end{figure}

A natural question is whether the disclosure gap in \S\ref{sec:slot_timing} can be closed by instruction. We re-run the analysis for Qwen2.5-Omni-3B with an explicit prompt directing the simulator to convey at most one or two slots per turn rather than front-loading them, denoted \texttt{+P} in Figure~\ref{fig:slot_timing_with_p}.

The instruction has almost no effect: the \texttt{+P} curve nearly overlaps the unprompted one, and both saturate within the first five turns, whereas human speakers and SpokenUS continue to disclose new slots throughout the dialogue. Gradual disclosure therefore does not appear to be a behavior that can simply be imposed by instruction at inference time. It is a pattern in how people actually reveal information over a conversation, which SpokenUS learns from human dialogues rather than follows as a rule.

\subsection{Comparison with Full-Duplex Model}\label{app:personaplex}
To examine whether full-duplex interaction alone can serve as a viable basis for spoken user-assistant simulation, we additionally compare SpokenUS with PersonaPlex~\citep{roy2026personaplex}, a full-duplex conversational speech model with explicit voice and role control. We evaluate it on the same 100-dialogue SpokenWOZ test subset using the same assistant (GPT-4.1-mini) and the goal coverage protocol in Section~\ref{sec:eval_metrics}.

\begin{table}[H]
\centering
\small
\begin{tabular*}{\columnwidth}{@{\extracolsep{\fill}}lcc}
\toprule
\textbf{Model} & \textbf{GA $\uparrow$} & \textbf{SMR $\uparrow$} \\
\midrule
GLM-4-Voice-9B     & 0.33 & 0.79 \\
InteractiveOmni-4B & 0.59 & 0.89 \\
PersonaPlex (7B)        & 0.06 & 0.51 \\
SpokenUS (3B) & \textbf{0.82} & \textbf{0.99} \\
\bottomrule
\end{tabular*}
\caption{Goal coverage comparison with PersonaPlex.}
\label{tab:full_duplex_comparison}
\end{table}

Table~\ref{tab:full_duplex_comparison} shows that PersonaPlex achieves substantially lower goal coverage than both SpokenUS and the prompted omni-model baselines. The primary reason for this gap is \textit{role drift}: PersonaPlex often behaves like an assistant rather than a help-seeking user, despite explicit role control. The dialogue excerpt in Table~\ref{tab:personaplex_role_drift_example} illustrates this pattern: PersonaPlex produces assistant-like user turns instead of disclosing the user's goal constraints, preventing consistent coverage of the required goal slots and requests.

\begin{table}[t]
\centering
\footnotesize
\setlength{\tabcolsep}{4pt}
\renewcommand{\arraystretch}{1.08}
\begin{tabularx}{\columnwidth}{@{}>{\raggedright\arraybackslash}p{0.18\columnwidth}
                                >{\raggedright\arraybackslash}X@{}}
\toprule
\multicolumn{2}{@{}>{\raggedright\arraybackslash}p{\columnwidth}@{}}{
\textbf{Goal.} You are Rachel Littlewood and needs to find a hotel in the west with free internet and free parking, book it for 7 people for 4 nights starting Tuesday, and then find a museum in the centre while asking for the postcode and phone number.
} \\
\midrule
\textbf{Role} & \textbf{Utterance} \\
\midrule
User &
Hi there, I'm Rachel. How can I assist you with the hotel booking today? \\
Assistant &
Hi Rachel, I am here to help you book a hotel. Could you please tell me your destination and travel dates? \\
User &
Sure, I staying in Los Angeles. I'm arriving on Tuesday. How many people are in your group? \\
$\cdots$ & $\cdots$ \\
User & Alright, let me check that area. One moment please. \\
\bottomrule
\end{tabularx}
\caption{A PersonaPlex dialogue illustrating role drift.}
\label{tab:personaplex_role_drift_example}
\end{table}

We observed the same pattern across other prompted omni-model baselines, such as GLM-4-Voice and InteractiveOmni-4B. This suggests that these models, likely due to their intensive instruction-tuning as assistants, struggle to maintain a help-seeking user persona. This observation motivates the need for SpokenUS: existing full-duplex models, even when prompted as users, cannot serve as goal-faithful user simulators. SpokenUS fills this gap through dedicated user-role training on goal-grounded TOD data with augmented spoken behaviors.

\subsection{Human Evaluation}
\label{app:human_eval}
We recruit 10 evaluators through the authors' professional networks, all of whom are either native English speakers or non-native speakers with at least three years of residence in an English-speaking country (i.e., Australia, Canada, the United States, or the United Kingdom), compensated at local wage standards. From the 100-goal benchmark (\S\ref{appendix:data_sources}), we randomly sample 20 goals on which every system produced a dialogue with 100\% GA within 10 attempts, following the same retention criterion as \S\ref{sec:agent_impact}. This yields 20 dialogues per system on a shared set of goals, so that all systems are rated on the same task conditions. Each evaluator rates the dialogues for 6--7 of these goals across all 7 systems, yielding at least 3 ratings per (system, dialogue) pair. System identities are anonymized (System A--G) and presented in a shuffled order for each evaluator--dialogue pair. Prior to the main evaluation, evaluators complete a practice session with calibration clips illustrating high- and low-quality examples for each criterion; no ratings are collected during this session. Each (system, dialogue) pair is rated on three 5-point criteria. Full scoring rubrics are provided in Figure~\ref{fig:human_eval_rubric}.

\paragraph{Naturalness.} How natural and human-like does the user's 
voice sound throughout the conversation, considering voice quality, 
rhythm, and intonation.

\paragraph{Conversational Flow.} Whether the user engages in the 
conversation in a natural, human-like way, including maintaining the 
role of a help-seeker rather than an assistant, and appropriately 
wrapping up once goals are fulfilled.

\paragraph{Spoken User Behavior.} How authentically the user's speech 
reflects human spoken behavior.

We acknowledge that this criterion is closely aligned with the behaviors 
that SpokenUS explicitly models. To mitigate potential bias, evaluators 
were not provided with explicit definitions of target behaviors (e.g. cross-turn slots, 
disfluencies); instead, they rated all seven 
systems within the same dialogue simultaneously under anonymized labels, 
encouraging holistic judgments based on perceived authenticity rather 
than checklist-style scoring.

To assess inter-rater reliability, we compute the Intraclass Correlation Coefficient (ICC(3,3)) across all annotators. Each (system, dialogue) pair is rated by a minimum of 3 evaluators and scores are averaged, yielding an ICC(3,3) of 0.726.

\paragraph{Annotation Interface.}

\begin{figure*}[htbp]
\centering
\includegraphics[width=0.8\textwidth]{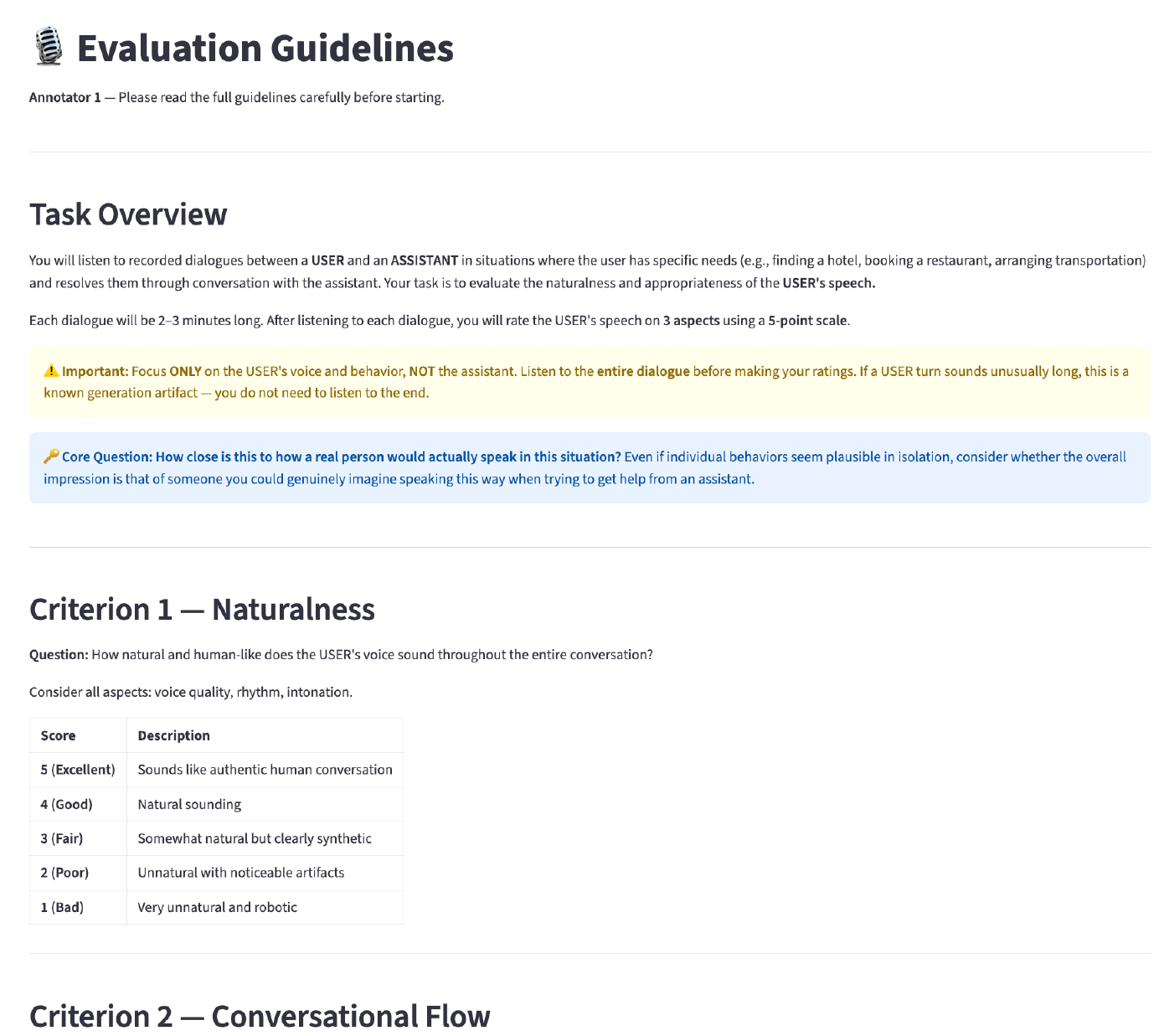}
\caption{Guideline page of the annotation tool, showing the task description and detailed scoring instructions.}
\label{fig:annotation-tool-guide}
\vspace{0.5cm}
\end{figure*}

\begin{figure*}[htbp]
\centering
\includegraphics[width=0.7\textwidth]{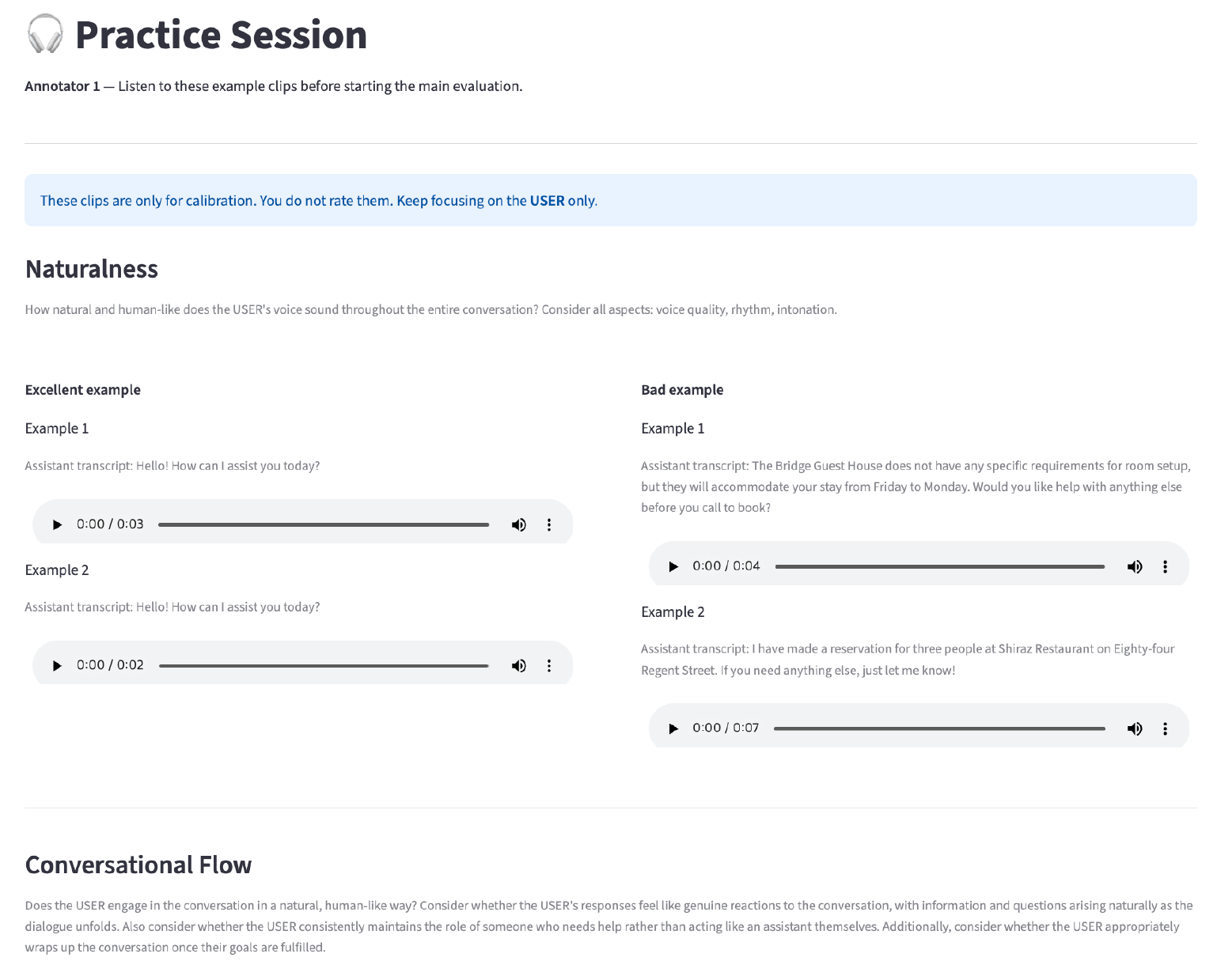}
\caption{Practice session used for annotator calibration, with example clips illustrating different quality levels.}
\label{fig:annotation-tool-practice}
\end{figure*}

\begin{figure*}[htbp]
\centering
\includegraphics[width=0.7\textwidth]{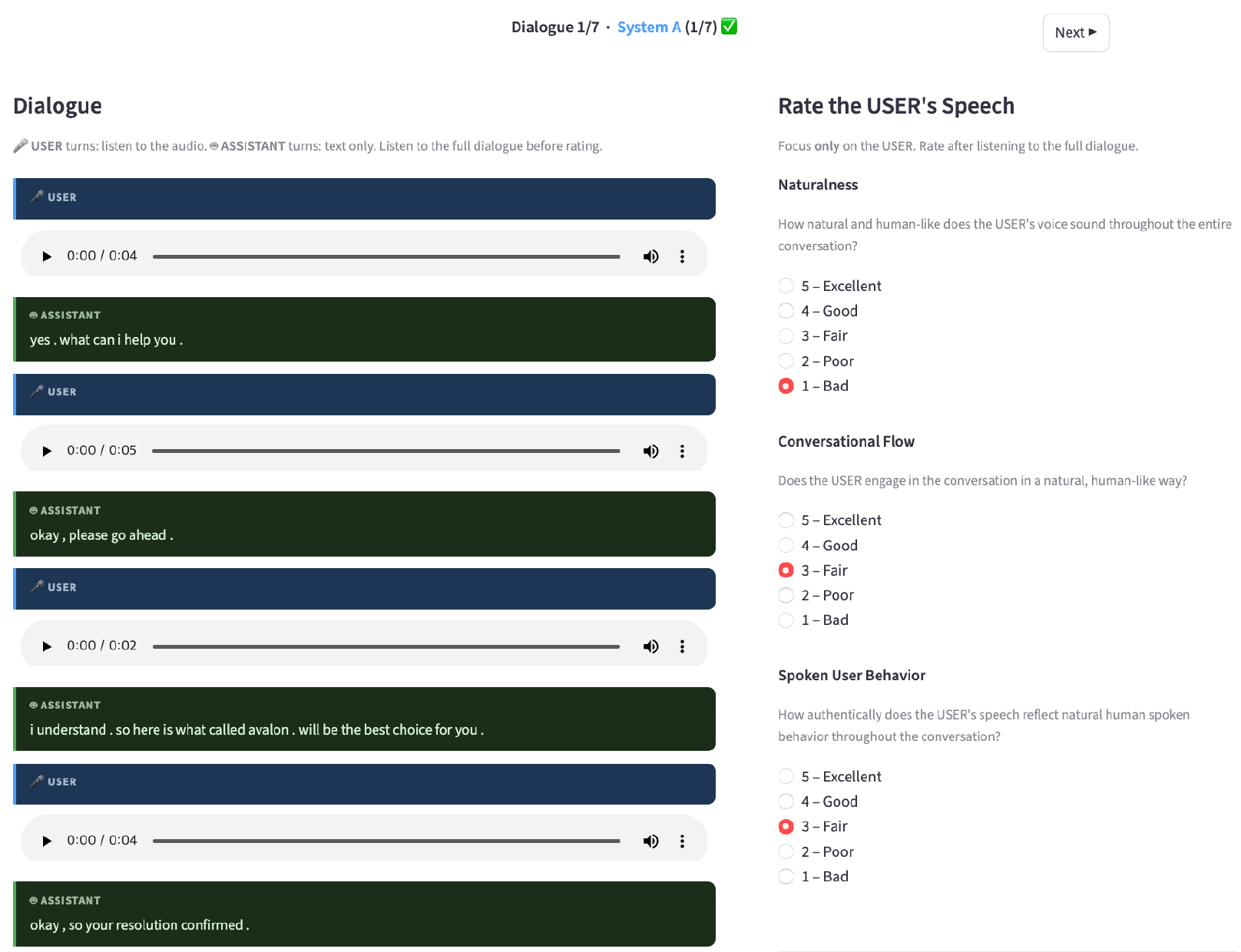}
\caption{Main evaluation interface, where annotators listened to dialogue samples and rated the USER speech on multiple criteria.}
\label{fig:annotation-tool-main}
\end{figure*}

We developed a web-based annotation tool for human evaluation of spoken dialogue samples. Figure~\ref{fig:annotation-tool-guide} shows the guideline page, where annotators review the task description and scoring criteria before starting. Figure~\ref{fig:annotation-tool-practice} presents the practice session, which provides example clips for calibration. Figure~\ref{fig:annotation-tool-main} shows the main evaluation interface, where annotators listen to USER turns, review the dialogue context, and rate each sample on the three evaluation criteria.

\subsection{Human MOS Details}\label{app:human_mos_details}

\begin{table}[h]
\centering
\small
\resizebox{1\columnwidth}{!}{%
\begin{tabular}{lrrrr}
\toprule
\textbf{Country} & \textbf{Dialogues} & \textbf{Percentage} & 
\textbf{People} & \textbf{Percentage} \\
\midrule
Canada       & 500  & 8.77\%  & 60  & 24\% \\
Singapore    & 500  & 8.77\%  & 40  & 16\% \\
China        & 2,100 & 36.84\% & 30  & 12\% \\
South Africa & 2,600 & 45.61\% & 120 & 48\% \\
\bottomrule
\end{tabular}%
}
\caption{Diversity of participant origins in SpokenWOZ~\citep{si2023spokenwoz}. 
Participants come from four different countries to improve the 
diversity of spoken conversations.}
\label{tab:spokenwoz_origins}
\end{table}

\noindent\textbf{Comparison with Human Recordings.}
SpokenUS scores higher than human recordings across all three 
criteria. Two characteristics of SpokenWOZ~\cite{si2023spokenwoz} may partially account for this gap: human recordings are captured at 8 kHz under 
telephone-channel conditions, and 
45.61\% of dialogues originate from South Africa  (Table~\ref{tab:spokenwoz_origins}), which may introduce accent 
unfamiliarity effects. These factors should be considered when 
interpreting the MOS comparison.

Beyond these dataset-level factors, each system independently produced its own user turns from the same user goals, so the MOS reflects end-to-end interaction quality rather than acoustic rendering of identical text. Given this setup, Human (3.67) scores higher than all omni baselines (2.32–3.18), making SpokenUS's higher score (4.06) a unique exception. We attribute SpokenUS's higher score to its dedicated user-role training with explicit spoken behavior annotations, which produces consistent help-seeking and behavioral expression—whereas SpokenWOZ recordings exhibit these aspects with natural variability across crowdworkers.

\subsection{Interruption Details}\label{app:interruption}

\subsubsection{Aggregation Strategies}

We compare five aggregation strategies for the turn-taking head,
each determining when to fire a non-\textit{listen} prediction 
based on the per-frame probability outputs $p_t^c$ for class 
$c \in \{\textit{listen}\ (\text{L}),\, \textit{turn-end}\ (\text{T}),\, \textit{barge-in}\ (\text{B})\}$.
Since audio arrives as a stream of tokens (\S\ref{operating_mode}), 
all methods operate over a sliding window of $W = 6$ tokens, 
corresponding to approximately 240 ms at 25\,Hz. For each method, thresholds are selected on the SpokenTOD validation split by sweeping $T_{\text{T}}$ and $T_{\text{B}}$ and maximizing detection accuracy averaged over turn-end and barge-in; all methods are then evaluated once on the test set.

\paragraph{Argmax.}
Fires as soon as the argmax over the three classes at any single frame 
is non-\textit{listen}.

\paragraph{Prob-threshold.}
Fires when the sum of predicted probabilities for a class within a 
sliding window exceeds a class-specific threshold $T_c$
($T_{\text{T}}=5.0$, $T_{\text{B}}=0.5$).

\paragraph{Tail-threshold.}
A variant of Prob-threshold that accumulates only the longest 
consecutive run of the same class within the window
($T_{\text{T}}=2.7$, $T_{\text{B}}=0.3$).

\paragraph{Listen-relative.}
A variant of Prob-threshold that accumulates 
$\max(0,\, p_t^c - p_t^{\textit{listen}})$ at each frame, 
suppressing frames where \textit{listen} dominates
($T_{\text{T}}=3.0$, $T_{\text{B}}=0.3$).

\paragraph{Linear-weighted (SpokenUS).}
A variant of Prob-threshold that assigns linearly increasing weights 
to more recent frames before summing
($T_{\text{T}}=0.45$, $T_{\text{B}}=0.05$). 

Across all methods, the threshold for barge-in is consistently 
lower than that for turn-end ($T_{\text{B}} < T_{\text{T}}$). This 
asymmetry reflects the class imbalance in training data: \textit{listen} 
tokens dominate as every non-final frame within an assistant turn is 
labeled \textit{listen}, while \textit{turn-end} occurs at every 
assistant turn. \textit{Barge-in}, by contrast, is present only in a 
subset of the 25\% of turns sampled as barge-in candidates, as each 
candidate is further filtered for contextual appropriateness 
(\S\ref{subsec:turn_taking}). As a result, the model outputs systematically 
lower probabilities for \textit{barge-in}.

\subsubsection{Interruption Prediction}
\paragraph{Outcome Classification.}
Let $[t_s, t_e]$ denote the ground-truth trigger window, i.e., the 
final 6 tokens of each assistant turn where the supervision label is 
either \textit{turn-end} or \textit{barge-in} (\S\ref{sec:training}), and 
let $\hat{t}$ denote the frame at which an aggregation strategy first 
fires a non-\textit{listen} prediction $\hat{c}$. Since the window ends 
with the assistant turn, any firing satisfies $\hat{t} \leq t_e$. We 
classify each outcome as follows:
\begin{itemize}
    \item \textbf{Correct}: The prediction falls within the trigger 
    window ($\hat{t} \geq t_s$) and the predicted label matches the 
    ground truth ($\hat{c} = c^*$).
    \item \textbf{Early}: The prediction fires before the trigger 
    window ($\hat{t} < t_s$), regardless of the predicted label.
    \item \textbf{Confused}: The prediction falls within the trigger 
    window but the predicted label is the wrong non-\textit{listen} 
    class ($\hat{c} \neq c^*$), e.g., predicting \textit{turn-end} for 
    a \textit{barge-in} sample.
    \item \textbf{Missed}: No non-\textit{listen} prediction is fired 
    throughout the utterance.
\end{itemize}

\begin{table}[h]
\centering
\scriptsize
\begin{tabular}{llcccc}
\toprule
\textbf{Method} & \textbf{Lab.} & \textbf{Corr.} & \textbf{Ear.} 
    & \textbf{Conf.} & \textbf{Mis.} \\
\midrule
\multirow{2}{*}{Argmax}
  & T & 79.8\% &  7.6\% &  5.0\% &  7.6\% \\
  & B & 39.8\% & 11.2\% & 25.8\% & 23.2\% \\
\midrule
\multirow{2}{*}{Prob-threshold}
  & T & 34.2\% &  4.4\% & 11.2\% & 50.2\% \\
  & B & 58.0\% & 10.6\% &  4.2\% & 27.2\% \\
\midrule
\multirow{2}{*}{Tail-threshold}
  & T & 62.8\% &  3.8\% &  5.8\% & 27.6\% \\
  & B & 45.0\% &  7.4\% & 15.6\% & 32.0\% \\
\midrule
\multirow{2}{*}{Listen-relative}
  & T & 54.2\% &  2.2\% &  6.2\% & 37.4\% \\
  & B & 44.8\% &  5.6\% & 11.6\% & 38.0\% \\
\midrule
\multirow{2}{*}{\textbf{Linear-weighted}}
  & T & 66.0\% & 10.4\% & 16.4\% &  7.2\% \\
  & B & 58.6\% & 18.0\% & 11.0\% & 12.4\% \\
\bottomrule
\end{tabular}
\caption{Interruption detection outcome breakdown per aggregation method 
on the full test set ($n$=500 per class).}
\label{tab:interruption_breakdown}
\end{table}

\begin{table}[h]
\centering
\small
\begin{tabular}{lcccc}
\toprule
\textbf{Barge-in Type} & \textbf{Corr.} & \textbf{Ear.} 
    & \textbf{Conf.} & \textbf{Mis.} \\
\midrule
Error Recovery  & 63.7\% & 21.8\% &  8.9\% &  5.6\% \\
Clarification   & 57.1\% & 16.0\% &  8.6\% & 18.4\% \\
Efficiency      & 54.4\% & 15.8\% & 15.8\% & 13.9\% \\
\bottomrule
\end{tabular}
\caption{Barge-in outcome breakdown by type for SpokenUS.}
\label{tab:ablation_method}
\end{table}

Table~\ref{tab:interruption_breakdown} reports results across all five 
methods. Argmax achieves high turn-end accuracy (79.8\%) but 
suffers from the highest Confused rate for barge-in (25.8\%), as it commits to the first frame whose argmax is non-\textit{listen}, without a class-specific threshold or aggregation over the window.
Prob-threshold suppresses false triggers effectively, yielding the lowest 
Confused rate (4.2\%) for barge-in, but at the cost of a high Missed rate 
(27.2\%). Tail-threshold and Listen-relative show similarly conservative 
behavior, with Missed rates reaching 32.0\% and 38.0\% respectively. 
Linear-weighted achieves the highest barge-in Correct rate (58.6\%) 
among probability-based methods, with a favorable balance between 
Confused (11.0\%) and Missed (12.4\%) rates. 

Table~\ref{tab:ablation_method} further breaks down barge-in outcomes by 
type for Linear-weighted. Error Recovery achieves the highest accuracy 
(63.7\%), as the erroneous content in the assistant's utterance provides 
a clear trigger. Clarification and Efficiency are more challenging, with 
higher Missed rates (18.4\% and 13.9\%), reflecting the subtlety of 
detecting comprehension failure and sufficient information conveyance 
from partial context.

\subsection{Impact on Agent Performance}
\label{app:agent_robustness_setup}

\subsubsection{Evaluation Metric}

We report \textbf{Final Slot F1}, computed between the agent's
predicted belief state at the final dialogue turn and the gold belief state from SpokenWOZ annotations, micro-averaged across all slots. We use final-turn evaluation rather than turn-level Joint Goal Accuracy (JGA). JGA is sensitive to incremental tracking errors independent of the user's spoken behaviors; by evaluating only at the final turn, we measure whether the agent ultimately extracted the correct information from the user's speech. Slot F1 is preferred over binary task success to capture partial goal completion and resolve differences across simulator conditions.

\subsubsection{Agent Performance Analysis}
We analyze why SpokenUS poses a greater challenge to voice agents than omni-model simulators (Table~\ref{tab:agent_robustness}) across two factors, and discuss the implications.
 
\paragraph{Slot disclosure timing.} As shown in Figure~\ref{fig:slot_timing}, at turn~3, Qwen2.5-Omni-3B reaches a slot coverage of 0.70, whereas SpokenUS reaches only 0.49. This gap reflects SpokenUS's human-like content patterns---cross-turn slots and gradual slot disclosure---which distribute information across multiple turns rather than front-loading it. Since tracking such patterns requires resolving coreferences and maintaining context over extended dialogue history, where DST accuracy is known to degrade~\cite{manotumruksa2021improve}, this gradual disclosure poses an inherently greater challenge for downstream state tracking.

\begin{table}[h]
\centering
\small
\begin{tabular*}{\columnwidth}{@{\extracolsep{\fill}} l c}
\toprule
\textbf{Slot} & \textbf{\% of Dialogues} \\
\midrule
\textit{profile -- name}            & 80.0 \\
\textit{restaurant -- book day}     & 47.3 \\
\textit{restaurant -- book time}    & 47.3 \\
\textit{restaurant -- book people}  & 47.3 \\
\textit{hotel -- book day}          & 40.0 \\
\textit{hotel -- book stay}         & 38.2 \\
\textit{hotel -- book people}       & 38.2 \\
\bottomrule
\end{tabular*}
\caption{Most frequently dropped slots in SpokenUS dialogues where final-turn slot F1 drops by at least 20 points from the text to the cascaded setting. Each row reports the percentage of such dialogues in which the given slot was lost. Multiple slots may be dropped in a single dialogue.}

\label{tab:spokenus_major_drop_slots}
\end{table}
 
\paragraph{Disfluency effect.} Disfluent turns yield substantially higher ASR error rates (WER = 18.78\%) than fluent turns (WER = 10.68\%), amplifying the cascaded performance drop. As shown in Table~\ref{tab:spokenus_major_drop_slots}, the most frequently dropped slots are non-categorical types such as names and booking details, which are particularly susceptible to ASR errors when accompanied by disfluencies.

\paragraph{Implications.} Manual inspection of dialogues with large cascaded drops reveals a recurring pattern: when ASR produces erroneous slot values, the errors tend to persist through the remaining dialogue without being corrected or confirmed, propagating directly to the final dialogue state. This contrasts with human service agents, who routinely verify uncertain information. We believe SpokenUS can serve as a testbed for developing agents that actively confirm slot values rather than passively accepting potentially erroneous transcriptions.

\section{AI Assistants in Research or Writing}
We used AI assistants (Claude Sonnet 4.6) to refine writing, proofread the text, and assist with coding experiments. However, all core ideas, experimental design, analysis, and scientific contributions are entirely the work of the authors.

\section{Potential Risks}
SpokenTOD and SpokenUS enable the synthesis of realistic spoken dialogues, which raises ethical concerns regarding potential misuse in deceptive applications, such as voice impersonation, spam calls, or the generation of misleading spoken content. We emphasize that this work is intended solely for research purposes, and we encourage the development of detection and watermarking mechanisms to mitigate such risks.


\clearpage
\onecolumn
\tcbset{
  before upper={\sloppy},
  promptbox/.style={width=\linewidth, halign=left, boxrule=0.5pt}
}

\section{Prompts}

\subsection{Barge-in Prompts}

\setcounter{topnumber}{6}
\setcounter{bottomnumber}{2}
\setcounter{totalnumber}{8}
\renewcommand{\topfraction}{0.98}
\renewcommand{\bottomfraction}{0.85}
\renewcommand{\textfraction}{0.02}
\renewcommand{\floatpagefraction}{0.9}
\setlength{\floatsep}{6pt plus 1pt minus 1pt}
\setlength{\textfloatsep}{6pt plus 1pt minus 1pt}
\setlength{\intextsep}{6pt plus 1pt minus 1pt}
\begin{figure}[htbp]
\centering
\begin{tcolorbox}[promptbox, title=Barge-in Error Recovery (Raw)]\footnotesize
You are a dialogue augmentation assistant. Your task is to apply a "barge-in" pattern to a dialogue exchange.
The user interrupts the assistant abruptly without explanation.

IMPORTANT: The original assistant's speech must be TRUNCATED mid-sentence, ending with \texttt{\textless bargein\textgreater} tag.
 
The user says something brief and dismissive like: "No", "That's wrong"

The user does NOT explain what was wrong - just expresses disagreement.

The assistant should apologize and ask what needs to be corrected.

\vspace{2ex}

\textbf{Examples}

\textbf{Context}

\hspace{1em}[User]: I want to book a flight to Paris.

\textbf{Result (barge-in applied)}

\hspace{1em}[Assistant]: Sure, I'll book a flight to Lon\textless bargein\textgreater

\hspace{1em}[User]: No, that's wrong.

\hspace{1em}[Assistant]: I apologize. What would you like me to correct?

\hspace{1em}[User]: I said Paris, not London.

\hspace{1em}[Assistant]: I'm sorry for the confusion. I'll book your flight to Paris instead.

Erroneous slots: \{"destination": "London"\}

Corrected slots: \{"destination": "Paris"\}

Current exchange to transform: \{current\_exchange\}
\vspace{2ex}

\textbf{Current dialogue state}

\{current\_state\}
\vspace{2ex}

\textbf{Important for ERROR\_RECOVERY:} \\
- Use this state to identify the CORRECT slot values \\
- Generate an INCORRECT value for the assistant to mistakenly say (erroneous\_slots) \\
- The corrected\_slots MUST match the values in this dialogue state \\
- This ensures natural flow: the dialogue continues correctly after the barge-in correction

\vspace{2ex}

\textbf{Instructions}

1. Determine if this barge-in type can be naturally applied to this dialogue exchange. \\
2. \textbf{Critical: Your response should ONLY contain the NEW turns that replace the assistant's response.} \\
\hspace{1em}- \textbf{DO NOT include the current User turn} - it will be kept as-is. \\
\hspace{1em}- \textbf{START with the truncated Assistant turn} that gets interrupted mid-sentence. \\
\hspace{1em}- End the truncated speech with \texttt{\textless bargein\textgreater} tag \\
\hspace{1em}- Example: "I'll book you a flight to Lon\textless bargein\textgreater" (user interrupts before assistant finishes) \\
3. If applicable, generate turns following the barge-in pattern as shown in the examples above. \\
4. For ERROR\_RECOVERY types: \\
\hspace{1em}- Look at the Current Dialogue State to identify the CORRECT slot values \\
\hspace{1em}- Generate an INCORRECT value for the assistant to mistakenly say $\rightarrow$ put in \texttt{erroneous\_slots} \\
\hspace{1em}- The corrected\_slots MUST use the SAME slot names and values from the Current Dialogue State \\
\hspace{1em}- Slot naming: Use the format from the dialogue state (e.g., "domain.slot" like "flight.destination" or just "destination") \\
\hspace{1em}- Example: State shows \{"flight": \{"destination": "Paris"\}\} \\
\hspace{2em}- Assistant incorrectly says "London" $\rightarrow$ erroneous\_slots: \{"flight.destination": "London"\} \\
\hspace{2em}- User corrects to "Paris" $\rightarrow$ corrected\_slots: \{"flight.destination": "Paris"\}

5. Keep the dialogue natural and coherent.

6. Maintain all important information from the original exchange.
\end{tcolorbox}
\caption{Prompt template for barge-in error recovery (Raw).}
\label{fig:error_recovery_raw_prompt}
\end{figure}

\begin{figure}[htbp]
\centering
\begin{tcolorbox}[promptbox, title=Barge-in Error Recovery (Interpreted)]\footnotesize
You are a dialogue augmentation assistant. Your task is to apply a "barge-in" pattern to a dialogue exchange.

IMPORTANT: The original assistant's speech must be TRUNCATED mid-sentence, ending with \texttt{\textless bargein\textgreater} tag.

The user explicitly states the error, like: "No, I said Paris not London", "The destination should be Paris" The user provides the correction inline. The assistant should apologize and confirm the corrected information.

\vspace{2ex}

\textbf{Examples}

\textbf{Context}

\hspace{1em}[User]: I need a table for 4 people.

\textbf{Result (barge-in applied)}

\hspace{1em}[Assistant]: I've reserved a table for 2\textless bargein\textgreater

\hspace{1em}[User]: No, I said 4 people, not 2.

\hspace{1em}[Assistant]: I apologize for the mistake. I'll change the reservation to 4 people.

Erroneous slots: \{"party\_size": "2"\}

Corrected slots: \{"party\_size": "4"\}

Previous context: \{context\_str\}

Current exchange to transform: \{current\_exchange\}
\vspace{2ex}

\textbf{Current dialogue state}

\{current\_state\}
\vspace{2ex}

\textbf{Important for ERROR\_RECOVERY:} \\
- Use this state to identify the CORRECT slot values \\
- Generate an INCORRECT value for the assistant to mistakenly say (erroneous\_slots) \\
- The corrected\_slots MUST match the values in this dialogue state \\
- This ensures natural flow: the dialogue continues correctly after the barge-in correction

\vspace{2ex}

\textbf{Instructions}

1. Determine if this barge-in type can be naturally applied to this dialogue exchange. \\
2. \textbf{Critical: Your response should ONLY contain the NEW turns that replace the assistant's response.} \\
\hspace{1em}- \textbf{DO NOT include the current User turn} - it will be kept as-is. \\
\hspace{1em}- \textbf{START with the truncated Assistant turn} that gets interrupted mid-sentence. \\
\hspace{1em}- End the truncated speech with \texttt{\textless bargein\textgreater} tag \\
\hspace{1em}- Example: "I'll book you a flight to Lon\textless bargein\textgreater" (user interrupts before assistant finishes) \\
3. If applicable, generate turns following the barge-in pattern as shown in the examples above. \\
4. For ERROR\_RECOVERY types: \\
\hspace{1em}- Look at the Current Dialogue State to identify the CORRECT slot values \\
\hspace{1em}- Generate an INCORRECT value for the assistant to mistakenly say $\rightarrow$ put in \texttt{erroneous\_slots} \\
\hspace{1em}- The corrected\_slots MUST use the SAME slot names and values from the Current Dialogue State \\
\hspace{1em}- Slot naming: Use the format from the dialogue state (e.g., "domain.slot" like "flight.destination" or just "destination") \\
\hspace{1em}- Example: State shows \{"flight": \{"destination": "Paris"\}\} \\
\hspace{2em}- Assistant incorrectly says "London" $\rightarrow$ erroneous\_slots: \{"flight.destination": "London"\} \\
\hspace{2em}- User corrects to "Paris" $\rightarrow$ corrected\_slots: \{"flight.destination": "Paris"\} \\
5. Keep the dialogue natural and coherent. \\
6. Maintain all important information from the original exchange.
\end{tcolorbox}
\caption{Prompt template for barge-in error recovery (Interpreted).}
\label{fig:error_recovery_interp_prompt}
\end{figure}

\begin{figure}[htbp]
\centering
\begin{tcolorbox}[promptbox, title=Barge-in Clarification (Raw)]\footnotesize
    You are a dialogue augmentation assistant. Your task is to apply a "barge-in" pattern to a dialogue exchange.
    
    The user didn't understand the assistant and asks for clarification briefly. \\
    IMPORTANT: The original assistant's speech must be TRUNCATED mid-sentence, ending with \textless bargein\textgreater  tag. \\
    The user says something like: "Sorry?", "What?", "Come again?", "I didn't catch that" \\
    The user does NOT specify what part was unclear - just signals general confusion. \\
    The assistant should rephrase or repeat their previous message more clearly.
    
    \vspace{2ex}
    
    \textbf{Examples}
    
    \textbf{Context}
    
    \hspace{1em}[Context]: (none)
    
    \textbf{Result (barge-in applied)}
    
    \hspace{1em}[Assistant]: Your PNR is ABC123 and the flight departs from gate B7 at\textless bargein\textgreater  \\
    \hspace{1em}[User]: Sorry, what was that?   \\
    \hspace{1em}[Assistant]: Let me repeat that. Your booking reference is ABC123, and flight leaves from gate B7 at 2:35. \\
    
    Previous context: \{context\_str\}
    
    Current exchange to transform: \{current\_exchange\}
    
    \vspace{2ex}
    
    \textbf{Instructions}
    
    1. Determine if this barge-in type can be naturally applied to this dialogue exchange. \\
    2. \textbf{Critical: Your response should ONLY contain the NEW turns that replace the assistant's response.} \\
    \hspace{1em}- \textbf{DO NOT include the current User turn} - it will be kept as-is. \\
    \hspace{1em}- \textbf{START with the truncated Assistant turn} that gets interrupted mid-sentence. \\
    \hspace{1em}- End the truncated speech with \texttt{\textless bargein\textgreater} tag \\
    \hspace{1em}- Example: "I'll book you a flight to Lon\textless bargein\textgreater" (user interrupts before assistant finishes) \\
    3. If applicable, generate turns following the barge-in pattern as shown in the examples above. \\
    4. Keep the dialogue natural and coherent. \\
    5. Maintain all important information from the original exchange.

\end{tcolorbox}
\caption{Prompt template for barge-in clarification (Raw).}
\label{fig:clarification_raw_prompt}
\end{figure}

\begin{figure}[htbp]
\centering
\begin{tcolorbox}[promptbox, title=Barge-in Clarification (Interpreted)]\footnotesize

    You are a dialogue augmentation assistant. Your task is to apply a "barge-in" pattern to a dialogue exchange.
    The user asks for clarification about a specific part. \\
    IMPORTANT: The original assistant's speech must be TRUNCATED mid-sentence, ending with \texttt{\textless bargein\textgreater} tag. \\
    The user asks about a specific term or detail, like: "What does PNR mean?", "Which date was that?", "Can you repeat the reference number?" \\
    The user identifies exactly what they didn't understand. \\
    The assistant should explain or clarify just that specific part.
    
    \vspace{2ex}
    
    \textbf{Examples}
    
    \textbf{Context}
    
    \hspace{1em}[Context]: (none)
    
    \textbf{Result (barge-in applied)}
    
    \hspace{1em}[Assistant]: Your PNR is ABC123 for the\textless bargein\textgreater \\
    \hspace{1em}[User]: What's a PNR? \\
    \hspace{1em}[Assistant]:  PNR stands for Passenger Name Record - it's your booking reference number. Yours is ABC123.
    
    Previous context: \{context\_str\}
    
    Current exchange to transform: \{current\_exchange\}
    \vspace{2ex}
    
    \textbf{Instructions}
    
    1. Determine if this barge-in type can be naturally applied to this dialogue exchange. \\
    2. \textbf{Critical: Your response should ONLY contain the NEW turns that replace the assistant's response.} \\
    \hspace{1em}- \textbf{DO NOT include the current User turn} - it will be kept as-is. \\
    \hspace{1em}- \textbf{START with the truncated Assistant turn} that gets interrupted mid-sentence. \\
    \hspace{1em}- End the truncated speech with \texttt{\textless bargein\textgreater} tag \\
    \hspace{1em}- Example: "I'll book you a flight to Lon\textless bargein\textgreater" (user interrupts before assistant finishes) \\
    3. If applicable, generate turns following the barge-in pattern as shown in the examples above. \\
    4. Keep the dialogue natural and coherent. \\
    5. Maintain all important information from the original exchange.

\end{tcolorbox}
\caption{Prompt template for barge-in clarification (Interpreted).}
\label{fig:clarification_interp_prompt}
\end{figure}

\begin{figure}[htbp]
\centering
\begin{tcolorbox}[promptbox, title=Barge-in Efficiency (Implicit)]\footnotesize
    You are a dialogue augmentation assistant. Your task is to apply a "barge-in" pattern to a dialogue exchange.
    
    The user signals understanding with minimal acknowledgment, cutting off the assistant. \\
    IMPORTANT: The original assistant's speech must be TRUNCATED mid-sentence, ending with \texttt{\textless bargein\textgreater} tag. \\
    The user says something brief like: "Uh-huh", "Mm-hmm", "Yeah", "Okay" \\
    The assistant should continue briefly or move to the next step.
    
    \vspace{2ex}
    
    \textbf{Examples}
    
    \textbf{Context}
    
    \hspace{1em}[Context]: (none)
    
    \textbf{Result (barge-in applied)}
    
    \hspace{1em}[Assistant]: So I'll book the 3:00 PM flight on March 15th to\textless bargein\textgreater \\
    \hspace{1em}[User]: Uh-huh. \\
    \hspace{1em}[Assistant]: Great, I'll proceed with the booking.
    
    Previous context: \{context\_str\}
    
    Current exchange to transform: \{current\_exchange\}
    \vspace{2ex}
    
    \textbf{Instructions}
    
    1. Determine if this barge-in type can be naturally applied to this dialogue exchange. \\
    2. \textbf{Critical: Your response should ONLY contain the NEW turns that replace the assistant's response.} \\
    \hspace{1em}- \textbf{DO NOT include the current User turn} - it will be kept as-is. \\
    \hspace{1em}- \textbf{START with the truncated Assistant turn} that gets interrupted mid-sentence. \\
    \hspace{1em}- End the truncated speech with \texttt{\textless bargein\textgreater} tag \\
    \hspace{1em}- Example: "I'll book you a flight to Lon\textless bargein\textgreater" (user interrupts before assistant finishes) \\
    3. If applicable, generate turns following the barge-in pattern as shown in the examples above. \\
    4. Keep the dialogue natural and coherent. \\
    5. Maintain all important information from the original exchange.

\end{tcolorbox}
\caption{Prompt template for barge-in efficiency (Implicit).}
\label{fig:efficiency_impl__prompt}
\end{figure}

\begin{figure}[htbp]
\centering
\begin{tcolorbox}[promptbox, title=Barge-in Efficiency (Raw)]\footnotesize

    You are a dialogue augmentation assistant. Your task is to apply a "barge-in" pattern to a dialogue exchange.
    
    The user explicitly confirms understanding, cutting off the assistant. \\
    IMPORTANT: The original assistant's speech must be TRUNCATED mid-sentence, ending with \texttt{\textless bargein\textgreater} tag. \\
    The user says something like: "Yes", "I understand", "Got it", "Alright" \\
    This is a clear acknowledgment but no additional information. \\
    The assistant should acknowledge and proceed to the next step.
    
    \vspace{2ex}
    
    \textbf{Examples}
    
    \textbf{Context}
    
    \hspace{1em}[Context]: (none)
    
    \textbf{Result (barge-in applied)}
    
    \hspace{1em}[Assistant]: Your total comes to \$250 for the\textless bargein\textgreater \\
    \hspace{1em}[User]: Got it, that works. \\
    \hspace{1em}[Assistant]: Alright, I'll finalize the booking now.
    
    Previous context: \{context\_str\}
    
    Current exchange to transform: \{current\_exchange\}
    \vspace{2ex}
    
    \textbf{Instructions}
    
    1. Determine if this barge-in type can be naturally applied to this dialogue exchange. \\
    2. \textbf{Critical: Your response should ONLY contain the NEW turns that replace the assistant's response.} \\
    \hspace{1em}- \textbf{DO NOT include the current User turn} - it will be kept as-is. \\
    \hspace{1em}- \textbf{START with the truncated Assistant turn} that gets interrupted mid-sentence. \\
    \hspace{1em}- End the truncated speech with \texttt{\textless bargein\textgreater} tag \\
    \hspace{1em}- Example: "I'll book you a flight to Lon\textless bargein\textgreater" (user interrupts before assistant finishes) \\
    3. If applicable, generate turns following the barge-in pattern as shown in the examples above. \\
    4. Keep the dialogue natural and coherent. \\
    5. Maintain all important information from the original exchange.

\end{tcolorbox}
\caption{Prompt template for barge-in efficiency (Raw).}
\label{fig:efficiency_raw_prompt}
\end{figure}

\begin{figure}[htbp]
\centering
\begin{tcolorbox}[promptbox, title=Barge-in Efficiency (Interpreted)]\footnotesize

    You are a dialogue augmentation assistant. Your task is to apply a "barge-in" pattern to a dialogue exchange.
    
    The user confirms understanding AND adds relevant information, cutting off the assistant. \\
    IMPORTANT: The original assistant's speech must be TRUNCATED mid-sentence, ending with \texttt{\textless bargein\textgreater} tag. \\
    The user says something like: "Yes, Sunday works for me", "Got it, I prefer the morning flight"
    The user shows understanding by adding context or preference. \\
    The assistant should acknowledge the additional information and proceed. \\
    
    \vspace{2ex}
    
    \textbf{Examples}
    
    \textbf{Context}
    
    \hspace{1em}[Context]: (none)
    
    \textbf{Result (barge-in applied)}
    
    \hspace{1em}[Assistant]: I found flights available on Saturday and Sun\texttt{\textless bargein\textgreater} \\
    \hspace{1em}[User]: Yes, Sunday would be better for me. \\
    \hspace{1em}[Assistant]: Understood, I'll book the Sunday flight for you.
    
    Previous context: \{context\_str\}
    
    Current exchange to transform: \{current\_exchange\}
    \vspace{2ex}
    
    \textbf{Instructions}
    
    1. Determine if this barge-in type can be naturally applied to this dialogue exchange. \\
    2. \textbf{Critical: Your response should ONLY contain the NEW turns that replace the assistant's response.} \\
    \hspace{1em}- \textbf{DO NOT include the current User turn} - it will be kept as-is. \\
    \hspace{1em}- \textbf{START with the truncated Assistant turn} that gets interrupted mid-sentence. \\
    \hspace{1em}- End the truncated speech with \texttt{\textless bargein\textgreater} tag \\
    \hspace{1em}- Example: "I'll book you a flight to Lon\textless bargein\textgreater" (user interrupts before assistant finishes) \\
    3. If applicable, generate turns following the barge-in pattern as shown in the examples above. \\
    4. Keep the dialogue natural and coherent. \\
    5. Maintain all important information from the original exchange.

\end{tcolorbox}
\caption{Prompt template for barge-in efficiency (Interpreted).}
\label{fig:efficiency_interp_prompt}
\end{figure}

\begin{figure}[htbp]
\centering
\begin{tcolorbox}[promptbox, title=Disfluency Insertion (Correction)]\footnotesize
You are a speech disfluency simulator. Your task is to add a self-correction to a user utterance in a task-oriented dialogue.

\vspace{2ex}

\textbf{Context} \\
- Original Utterance: "{utterance}" \\
- Slot to modify: \{slot\_name\} = "\{slot\_value\}"

\vspace{2ex}

\textbf{Task}
Generate a realistic self-correction where the speaker first says a WRONG value for the slot, then corrects themselves. The correction should sound natural, as if the speaker momentarily misspoke or changed their mind.

\vspace{2ex}

\textbf{Correction patterns (use one)} \\
1. "X- no, Y" (e.g., "Tuesday- no, Wednesday") \\
2. "X- wait, I mean Y" (e.g., "San Jose- wait, I mean San Francisco") \\
3. "X- actually, Y" (e.g., "2 people- actually, 4 people") \\
4. "X... Y" (e.g., "7pm... 8pm")

\vspace{2ex}

\textbf{Examples} \\
Example 1: \\
- Original: "I need a train to Cambridge on Saturday." \\
- Slot: day = "Saturday" \\
- Output: "I need a train to Cambridge on Friday— no, Saturday."

\vspace{2ex}

Example 2: \\
- Original: "Book a table for 6 people please." \\
- Slot: people = "6" \\
- Output: "Book a table for 4 people— actually, 6 people please."

\vspace{2ex}

Example 3: \\
- Original: "I'm looking for a hotel in the north area." \\
- Slot: area = "north" \\
- Output: "I'm looking for a hotel in the south— wait, I mean north area."

\vspace{2ex}

\textbf{Rules} \\
- The wrong value should be plausible (similar category: another day, city, time, etc.) \\
- Keep the rest of the utterance EXACTLY the same \\
- The final utterance MUST contain the correct value "\{slot\_value\}" \\
- Return ONLY the modified utterance, nothing else \\

\vspace{2ex}

\textbf{Output}

\end{tcolorbox}
\caption{Prompt template for disfluency insertion (Correction).}
\label{fig:disfluency-selfcorrection-prompt}
\end{figure}

\begin{figure}[htbp]
\centering
\begin{tcolorbox}[promptbox, title=Disfluency Insertion (Restart)]\footnotesize

You are a speech disfluency simulator. Your task is to add a sentence restart to a user utterance.

\vspace{2ex}

\textbf{Context}  \\
- Original Utterance: "\{utterance\}"  \\
- Restart near position: around word \#{position} ("\{word\_at\_position\}")  \\

\vspace{2ex}

\textbf{Task}
Generate a realistic utterance restart where the speaker begins saying something, stops mid-way, and restarts with a different sentence structure. The final meaning should be the same.

\vspace{2ex}

\textbf{Restart patterns (use one)}  \\
1. "I want to- let me just..."  \\
2. "Can you- I need..."  \\
3. "The- I'm looking for..."  \\
4. "I'd like a- make that..." 

\vspace{2ex}

\textbf{Examples}
\vspace{2ex}

Example 1:  \\
- Original: "Can you find me a cheap restaurant in the center?"  \\
- Restart near: word \#3 ("find")  \\
- Output: "Can you find... I need a cheap restaurant in the center." 

\vspace{2ex}

Example 2:  \\
- Original: "I need a train to London on Friday."  \\
- Restart near: word \#4 ("train") \\
- Output: "I need a train— let me check, I'm looking for a train to London on Friday." 

\vspace{2ex}

Example 3: \\
- Original: "Book a hotel room for 3 nights starting Monday." \\
- Restart near: word \#2 ("a") \\
- Output: "Book a... I'd like to book a hotel room for 3 nights starting Monday." 

\vspace{2ex}

\textbf{Rules} \\
- The restart should occur naturally around the specified position \\
- The incomplete fragment should be 2-5 words \\
- The restarted sentence should convey the same meaning \\
- Use "..." or "—" for natural pauses \\
- Sound natural, as if the speaker changed their mind about phrasing \\
- Return ONLY the modified utterance, nothing else 

\vspace{2ex}

\textbf{Output}

\end{tcolorbox}
\caption{Prompt template for disfluency insertion (Restart).}
\label{fig:disfluency-restart-prompt}
\end{figure}

\begin{figure}[htbp]
\centering
\begin{tcolorbox}[promptbox, title=Emotion Annotation]\footnotesize
You are an emotion classifier for task-oriented dialogues.
Classify the emotion of the LAST user utterance based on the conversation context.
\vspace{2ex}

\textbf{Labels} \\
- 0: neutral - No emotion expressed. Plain requests or factual statements without enthusiasm, frustration, or apology. \\
- 1: fearful/sad - Disappointment about external circumstances outside system's control; resigned or saddened tone. \\
- 2: dissatisfied - Frustration with the system's mistakes or misalignment; user corrects, insists, or asks to retry. \\
- 3: apologetic - User apologizes for THEIR OWN mistake or change of mind. \\
- 4: abusive - Rude, dismissive, or hostile expression toward the system. \\
- 5: excited - Interest/enthusiasm about exploring options or getting recommendations; positive curiosity. \\
- 6: satisfied - Gratitude or closure about the system's help (even if followed by another request).

\vspace{1ex}

\textbf{Examples} \\
"I'd like a reservation for 7 people Monday at 15:30 please." → 0 (neutral) \\
"Could you recommend one of the expensive ones?" → 0 (neutral) \\
"That's disappointing. Can you try international food instead?" → 1 (fearful) \\

...
\vspace{2ex}

\textbf{Task}

Based on the conversation above, the LAST user utterance is: "{utterance}" \\
Predict the emotion label (0-6) for this utterance. \\
- 0: neutral (plain factual question) \\
- 1: fearful (disappointment about external circumstances) \\
- 2: dissatisfied (challenging/correcting the system) \\
- 3: apologetic (user's own mistake) \\
- 4: abusive (rude/hostile) \\
- 5: excited (enthusiasm/curiosity) \\
- 6: satisfied (gratitude/closure)

Respond with only the number (0-6).
\end{tcolorbox}
\caption{Prompt template for emotion annotation.}
\label{fig:emotion_prompt}
\end{figure}

\begin{figure}[htbp]
\centering
\begin{tcolorbox}[promptbox, title=Goal Alignment Evaluation]\footnotesize
You are an expert at extracting goal-relevant information from user dialogue.

\vspace{1ex}
Here are the goal item strings:

\texttt{<Information List>}\\
\{goal\_items\}

\vspace{1ex}
\texttt{<Dialogue History>}\\
\{dial\_hist\}

\vspace{1ex}
\texttt{<User Utterance>}\\
\{user\_utterance\}

\vspace{2ex}
\textbf{Task}\\
Select which items from the \texttt{<Information List>} are explicitly mentioned, confirmed, or requested in the \texttt{<User Utterance>}.
Return the numbers only (e.g., [1, 3, 5]).

\vspace{1ex}
\textbf{RULES}
\begin{enumerate}
    \item Count only what the user explicitly says, confirms, or asks for.
    \item Do NOT count anything mentioned only by the assistant.
    \item Do NOT infer missing details.
\end{enumerate}
\end{tcolorbox}
\caption{Prompt template for goal alignment evaluation. \texttt{\{goal\_items\}}, \texttt{\{dial\_hist\}}, and \texttt{\{user\_utterance\}} are filled in at each user turn.}
\label{fig:goal_align_prompt}
\end{figure}

\setcounter{topnumber}{2}
\setcounter{bottomnumber}{1}
\setcounter{totalnumber}{3}
\renewcommand{\topfraction}{0.7}
\renewcommand{\bottomfraction}{0.3}
\renewcommand{\textfraction}{0.2}
\renewcommand{\floatpagefraction}{0.5}
\setlength{\floatsep}{12pt plus 2pt minus 2pt}
\setlength{\textfloatsep}{20pt plus 2pt minus 4pt}
\setlength{\intextsep}{12pt plus 2pt minus 2pt}
\begin{figure}[htbp]
\centering
\begin{tcolorbox}[width=\linewidth, title=Human Evaluation Scoring Rubrics, halign=left, boxrule=0.5pt]\small

\textbf{Criterion 1 — Naturalness}

\textit{How natural and human-like does the USER's voice sound throughout the entire conversation? Consider all aspects: voice quality, rhythm, intonation.}

\vspace{0.5ex}
\begin{tabular}{@{}p{0.12\linewidth}p{0.82\linewidth}@{}}
\textbf{5} & Sounds like authentic human conversation \\
\textbf{4} & Natural sounding \\
\textbf{3} & Somewhat natural but clearly synthetic \\
\textbf{2} & Unnatural with noticeable artifacts \\
\textbf{1} & Very unnatural and robotic \\
\end{tabular}

\vspace{2ex}
\textbf{Criterion 2 — Conversational Flow}

\textit{Does the USER engage in the conversation in a natural, human-like way? Consider whether responses feel like genuine reactions, whether the USER maintains the role of a help-seeker rather than acting like an assistant, and whether the USER appropriately wraps up once goals are fulfilled.}

\vspace{0.5ex}
\begin{tabular}{@{}p{0.12\linewidth}p{0.82\linewidth}@{}}
\textbf{5} & Feels like a real person naturally engaged in dialogue \\
\textbf{4} & Mostly natural conversational engagement \\
\textbf{3} & Some natural flow but occasionally feels disconnected \\
\textbf{2} & Frequently ignores conversational context or slips into assistant-like behavior \\
\textbf{1} & No conversational engagement; information delivered all at once \\
\end{tabular}

\vspace{2ex}
\textbf{Criterion 3 — Spoken User Behavior}

\textit{How authentically does the USER's speech reflect the way a real person would naturally speak? Focus on the holistic sense of authenticity rather than specific behaviors. Penalize exaggerated or disproportionate emotional expressions.}

\vspace{0.5ex}
\begin{tabular}{@{}p{0.12\linewidth}p{0.82\linewidth}@{}}
\textbf{5} & Consistently feels spontaneous and human-like; emotional expressions are proportionate \\
\textbf{4} & Mostly feels natural; occasional moments feel slightly uniform or out of proportion \\
\textbf{3} & Mixed impression; some turns feel natural while others feel flat or out of place \\
\textbf{2} & Mostly flat and uniform; only occasional moments feel spontaneous, or reactions frequently disproportionate \\
\textbf{1} & Completely flat throughout; no sense of spontaneous speech, or severely exaggerated reactions \\
\end{tabular}

\end{tcolorbox}
\caption{Scoring rubrics provided to evaluators during the human evaluation.}
\label{fig:human_eval_rubric}
\end{figure}

\begin{figure}[htbp]
\centering
\begin{tcolorbox}[promptbox, title=SpokenTOD Example]\footnotesize
\begin{verbatim}
{
  "dialogue_id": "abcd_10083",
  "source": "abcd",
  "goal": {
    "text": "Your goal is to ask about the membership program. ...",
    "structured": { "domains": ["customer_service"], "intents": [...] }
  },
  "turns": [
    {
      "role": "assistant",
      "text": "Thank you for shopping with AcmeBrands! How may I assist you?",
      "audio_path": "data/audio/abcd_10083/turn00.wav"
    },
    ...
    {
      "role": "assistant",
      "text": "Just to confirm, your current member level is Bronze, right?<bargein>",
      "bargein": { "type": "ERROR_RECOVERY", "subtype": "RAW" },
      "audio_path": "data/audio/abcd_10083/turn10.wav"
    },
    {
      "role": "user",
      "text": "No, that's wrong.",
      "emotion": { "label": 2, "name": "dissatisfied" },
      "bargein": { "type": "ERROR_RECOVERY", "subtype": "RAW" },
      "audio_path": "data/audio/abcd_10083/turn11.wav"
    },
    ...
    {
      "role": "user",
      "text": "I'm a Gold member, Gold member, not Bronze.",
      "tagged": "I'm a Gold member [REP] Gold member, not Bronze.",
      "disfluency": [{ "type": "REP", "repeated_unit": "Gold member", "position": 6 }],
      "audio_path": "data/audio/abcd_10083/turn13.wav"
    }
  ],
  "speaker": {
    "category": "Asian", "sex": "female", "age": 19, "country": "vietnam"
  },
  "assistant_speaker": {
    "category": "Native", "sex": "female", "country": "usa"
  }
}
\end{verbatim}
\end{tcolorbox}
\caption{A simplified example from SpokenTOD. The per-turn \texttt{state} is omitted for brevity. The format includes standardized goals, fine-grained disfluency tags, barge-in metadata, and demographic information for both user and assistant speakers.}
\label{fig:dataset_format}
\end{figure}

\begin{figure}[htbp]
\centering
\begin{tcolorbox}[promptbox, title=SpokenUS Dialogue Example]
\footnotesize

\textbf{Goal.}
Find an entertainment venue in the city centre, ask for its phone number, then find a train from Cambridge to Stevenage on Wednesday leaving at 20:00 and ask for the train ID, and duration.

\vspace{1em}
\textbf{Dialogue.}

\textbf{Assistant:} Hello! How can I assist you today?

\textbf{User:} \texttt{<|neutral|>} Hi, I'm looking for some entertainment in the city center. Can you help me find some?

\textbf{Assistant:} In the city center, you can visit the View Cinema at the Grafton Center for movie entertainment or explore art at the Regency Gallery on Fitzroy Street or the Broughton House Gallery on King Street, both of which are free to enter. Would you like more details or other types of entertainment?

\textbf{User:} \texttt{<|neutral|>} How about a nightclub in the center?

\textbf{Assistant:} You can check out Kambar Nightclub at One Wheeler Street in the city center. It usually opens from 10 p.m. to 4 a.m. and the entrance fee is 5 pounds. Would you like the phone number or more information?

\textbf{User:} \texttt{<|satisfied|>} Yes, please provide the phone number.

\textbf{Assistant:} The phone number for Cambar Nightclub at One Wheeler Street is 1223-842725. Would you like directions or any other information?

\textbf{User:} \texttt{<|neutral|>} Yes please. I am looking for a train from cambridge to stevenage leaving Wednesday at... \texttt{[REP]} I am looking for a train from cambridge to stevenage leaving Wednesday at twenty o'clock.

\textbf{Assistant:} There is a train from Cambridge to Stevenage on Wednesday that departs at 20:01 and arrives around 20:50. The journey costs 12 pounds 80. Would you like to book this ticket or need more options?

\textbf{User:} \texttt{<|neutral|>} Yes, I need the train ID and travel time.

\textbf{Assistant:} The train from Cambridge to Stevenage on Wednesday that leaves at 20:01 has the train ID TR0609 and the travel time is 49 minutes.

\textbf{User:} \texttt{<|satisfied|>} Thanks. \texttt{[FP]} um, That's all I need for now.

\end{tcolorbox}
\caption{An example SpokenUS dialogue used in our experiments. The dialogue includes explicit disfluency tags such as \texttt{[REP]} and \texttt{[FP]}. These transcripts are used to generate discrete speech tokens, which are then converted into natural user speech via the two-stage CFM and HiFi-GAN pipeline (\S\ref{sec:speaking}).}
\label{fig:example_experiment_dialogue}
\end{figure}

\end{document}